\RequirePackage{fix-cm}
\documentclass[smallextended,natbib]{svjour3}       
\smartqed  
\usepackage{graphicx}
\usepackage{multicol,multirow}
\usepackage{lscape}
\usepackage{adjustbox}
\usepackage{rotating}
\usepackage{subcaption}
\usepackage{array}

\makeatletter
\renewcommand\paragraph{\@startsection{paragraph}{4}{\z@}%
                                     {-3.25ex\@plus -1ex \@minus -.2ex}%
                                     {1.5ex \@plus .2ex}%
                                     {\emph\normalfont\normalsize}}
\makeatother

\journalname{International Journal of Artificial Intelligence in Education}
\usepackage[dont-mess-around]{fnpct}
\usepackage{etoolbox}
\newcommand*{\affmark}[1][*]{\textsuperscript{#1}}

\begin{document}

\title{Automated Speech Scoring System Under The Lens
}
\subtitle{Evaluating and interpreting the linguistic cues for language proficiency\protect\affmark[1]\thanks{\affmark[1]Accepted for publication in the International Journal of Artificial Intelligence in Education (IJAIED)}}


\author{Pakhi Bamdev\affmark[2]\thanks{\affmark[2]Equal Contribution.} \and
        Manraj Singh Grover\affmark[2] \and
        Yaman Kumar Singla\affmark[2] \and
        Payman Vafaee \and
        Mika Hama \and
        Rajiv Ratn Shah
}





\institute{Pakhi Bamdev \at
           MIDAS Lab, Indraprastha Institute of Information Technology, Delhi, India \\
              \email{pakhii@iiitd.ac.in} \and
          Manraj Singh Grover \at
          MIDAS Lab, Indraprastha Institute of Information Technology, Delhi, India \\
              \email{manrajg@iiitd.ac.in, manrajsinghgrover@gmail.com} \and  
          Yaman Kumar Singla \at
          IIIT-Delhi, SUNY at Buffalo, and Adobe Media Data Science Research \\
              \email{yamank@iiitd.ac.in, ykumar@adobe.com} \and
          Payman Vafaee \at
          Columbia University, New York, United States of America \\
              \email{pv2203@tc.columbia.edu} \and
          Mika Hama \at
          Second Language Testing, Inc., United States of America \\
              \email{mhama@2lti.com} \and
          Rajiv Ratn Shah \at
          MIDAS Lab, Indraprastha Institute of Information Technology, Delhi, India \\
              \email{rajivratn@iiitd.ac.in}
}

\date{Received: date / Accepted: date}

\maketitle

\begin{abstract}

English proficiency assessments have become a necessary metric for filtering and selecting prospective candidates for both academia and industry. With the rise in demand for such assessments, it has become increasingly necessary to have the automated human-interpretable results to prevent inconsistencies and ensure meaningful feedback to the second language learners. Feature-based classical approaches have been more interpretable in understanding what the scoring model learns. Therefore, in this work, we utilize classical machine learning models to formulate a speech scoring task as both a classification and a regression problem, followed by a thorough study to interpret and study the relation between the linguistic cues and the English proficiency level of the speaker. First, we extract linguist features under five categories (fluency, pronunciation, content, grammar and vocabulary, and acoustic) and train models to grade responses. In comparison, we find that the regression-based models perform equivalent to or better than the classification approach. Second, we perform ablation studies to understand the impact of each of the feature and feature categories on the performance of proficiency grading. Further, to understand individual feature contributions,  we present the importance of top features on the best performing algorithm for the grading task. Third, we make use of Partial Dependence Plots and Shapley values to explore feature importance and conclude that the best performing trained model learns the underlying rubrics used for grading the dataset used in this study.

\end{abstract}

\section{Introduction}
\label{sec:introduction}

The English language proficiency assessments are designed to evaluate the ability of non-native language speakers to produce a spontaneous response to a given prompt. These responses are then evaluated by human experts based on various scoring rubrics and a score is given on a standard scale. These accredited scores often support both academia as well as the industry in making important decisions like admission to schools, hiring prospective employees, etc. Given the scale of such assessments\footnote{Each year, over 2.3 million people give the TOEFL exam.}, over the last two decades, there has been a growing interest \citep{ijcai2019-879,speechrater} in automating these proficiency assessments for second language learners, both in text-based \citep{burstein-chodorow-1999-automated,DIKLI20141} and speech-based evaluations  \citep{speechrater,8462562}. Automating the scoring of these responses alleviates the involved human inconsistencies. It also enables scoring and providing feedback at scale reducing the human effort and the cost involved.

The general practice in building such an automated scoring system involves collecting a corpus of responses that are scored by expert human raters followed by the feature extraction process and using machine learning to learn a model that map features to the human scores. This machine learning model is then used to estimate scores for unseen responses. For essay scoring systems \citep{ijcai2019-879}, in addition to extracted linguistic features, the input can include features from the prompt shown to the examinee, a knowledge base related to the prompt, other responses to the same prompt, etc. In the case of oral proficiency scoring, these systems \citep{speechrater} also use speech and prosody features extracted from audio response through the alignment process. Automatic Speech Scoring has been widely studied and worked upon both by academia as well as the industry that markets speech scoring systems and assessments. While the advancement in machine learning allows such scoring systems to work really well and produce scores that have good agreement with the human raters, they do not provide much interpretation to explain how the input features were analyzed. Most of the test takers appearing for English proficiency tests--- speech or text-based--- use these scores to apply for either a job position or a university for higher education. The lack of human interpretable interpretations on how the automated versions of language proficiency work limit the test taker's ability to improve on the right aspects of their speaking abilities. Therefore, interpretation of machine learning models is much needed not only to increase the user's faith in the automated system but to gain useful insights to improve products, processes, and research. Making a model interpretable might help in building understanding as to what the model observes while making a particular prediction and it can be used to formulate constructive feedback for the test takers to improve upon. Thus, in this paper, we depend heavily upon hand-crafted linguistic features across all the major domains of speech proficiency--- fluency, suprasegmental pronunciation, grammar and vocabulary, and content. We also include a few acoustic features to see the evolution of response over time. We model both regression and classification formulation of the speech scoring task, select the best performing approach, and try to employ the use of model-agnostic methods to interpret the selected machine learning model. We also establish a relationship between features and their impact on the model's scoring ability while revealing feature values and ranges that might lead to higher or lower scores.

\begin{figure}[thp]
    \centering
    \includegraphics[width=\columnwidth]{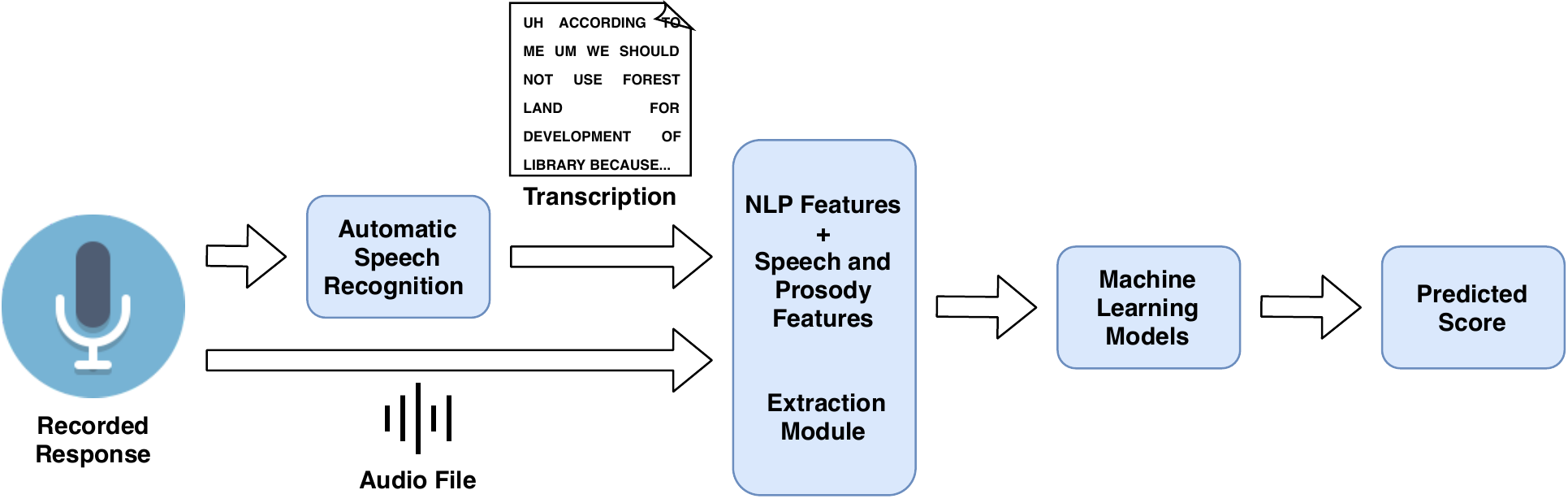}
    \caption{Pipeline for our automated speech scoring system.}
    \label{fig:architecture}
\end{figure}

The pipeline for our non-native automatic speech scoring system (see Figure~\ref{fig:architecture}) is as follows. The monologue audio response is passed through an Automatic Speech Recognition (ASR) system to generate its transcription. This ASR system is pre-trained on public datasets and later on in-domain responses for improved transcription accuracy. The combination of transcription and raw audio is then passed through a feature extractor module that aligns spoken words on an audio timeline and extracts linguistic features. The features are then used for training and evaluating various machine learning models for oral proficiency scoring.

Further, we perform analysis in order to understand what the top-performing model understands and correlate it with linguistic cues used during the rating process. Concretely, we report feature importance for top discriminating features and perform an ablation study to understand the contribution of features under each linguistic category. To understand how the variation of feature value impacts oral proficiency score, we utilize Partial Dependence Plots \citep{pdpth} and Shapley values \citep{shapth} to reveal the influence of the few top selected features on the model's ability to score. The main contribution of the paper is to make the automatic speech scoring model interpretable. We draw visualizations for effective communication of the insights on the proficiency scores obtained by the L2 Filipino English speakers. 

The rest of the paper is organized as follows. Section~\ref{sec:related_work} discusses the previous literature on automated scoring and interpretability in detail. Section~\ref{sec:dataset} describes the dataset used for training oral proficiency scoring, scoring rubric and analysis. Section~\ref{sec:features} elaborates on the ASR trained, transcription alignment process, and the features and their extraction process. Section~\ref{sec:experimental_setup} explains the experimental setup and metrics to evaluate the performance of the trained models. Section~\ref{sec:results} shares the model performance on unseen data, evaluates the impact of dropping individual feature category on scoring performance and further analyzes the top model through Partial Dependence Plots and Shapley values. Finally, Section~\ref{sec:conclusion} shares conclusion remarks and possible directions for future work.

\section{Related Work}
\label{sec:related_work}


The research on automated scoring systems can be dated to the year 1966 when \cite{page1966imminence} started scoring essays using punch cards. The advancements in technology and machine learning lead to more research exploring the scoring of specific aspects of responses as well as the overall response, both for text and speech modalities. In the case of speech modality, most oral proficiency systems use a manually engineered set of features, extracted using ASR hypothesis and prosodic analysis for grading spoken responses \citep{speechrater,aspiringminds}. These features pertain to various theoretical aspects of evaluating oral proficiency, including (but not limited to), pronunciation, fluency, content, vocabulary, and grammar, acoustic, etc.


One such comprehensive automated scoring system is SpeechRater\textit{\textsuperscript{SM}} \citep{speechrater} which extracts a rich set of acoustic and linguistic features for modeling and analysis purposes. The system combines years of research done by Educational Testing Service (ETS) in language proficiency automation. In this paper, we provide side by side comparison of correlation between features and target extracted by ETS and our system. \cite{loukina-etal-2015-feature} explored feature selection methodologies for automated speech scoring systems that satisfy the requirement of validity and interpretability while maintaining performance.
\cite{loukina-etal-2017-speech} also extracted acoustic and linguistic feature categories using ASR transcription and audio response, and combined them in combinations to understand impact of individual features in performance for oral proficiency task. \cite{aspiringminds} studied impact of errors generated by ASR on scoring performance and proposed a crowdsource way of improving transcription accuracy, and thereby, speech scoring performance. Recently, \cite{yoon-etal-2018-word} proposed features to score content of the response. They combined the content features with previous linguistic features and showed small but significant improvement in system's performance. 


The recent advancements in deep learning and improvement in automatic speech recognition's accuracy have enabled the development of end-to-end systems that can improve the overall performance of scoring models while alleviating the need to perform the complex tasks of manually engineering features and its pipeline. For example, the work done by \cite{8462562} proposed encoding lexical and acoustic cues using Bi-directional LSTM with Attention (BLSTM-Attention) neural networks to learn predictive features for scoring oral test responses. Their results showed a significant improvement in the scoring results over the manually-extracted feature-based scoring models. \cite{craighead-etal-2020-investigating} explored text-based auxiliary tasks and train models in a multi-task manner using speech transcription and found the L1 prediction task to benefit the scoring performance. Recent work also explored specific aspects of speech scoring like response content scoring, where, the features from the transcription of response are modeled with a respective question to learn the relevance of response \citep{yoon-lee-2019-content,8639697}. \cite{8683717} build over the work done by \cite{8639697} and model acoustic cues, prompt, and grammar features to improve scoring performance. \citet{singla2021speaker} in a recent work, use speech and text transformers \citep{shah2021all} to score candidate speech. They present an interesting analysis in which they extract cross-modal information from a hierarchy of responses and improve the scoring performance. 

Automatic scoring is used by millions of candidates for life-changing decisions including visas, college admissions, job interviews, \textit{etc.} Therefore, another pertinent and interesting related task is testing and improving the robustness of automatic scoring systems. In a series of studies, \citet{kumar2020calling,ding2020don,parekh2020my,singla2021aes} showed that automatic essay scoring systems are both overstable (significant changes in the essay text bring about little changes in the essay score) and oversensitive (little changes in the essay text bring about major changes in the essay score). They explored the impact of dataset vulnerabilities behind the adversarial susceptibility of the models. Further, they provided algorithms to resolve the oversensitivity and overstability of the deep learning models \citep{singla2021aes}. \citet{Malinin2017,raina-etal-2020-complementary} provide models to filter off-topic responses to improve the robustness of the system. This involves classifying irrelevant responses before passing it through the scoring pipeline. In an orthogonal method for improving institutional trust in automatic scoring systems, \citet{singla2021using} use sampling based methods to estimate and improve the accuracy of scoring models. Using their proposed reward sampling, they show significant improvements in model performance while using limited human scoring cost.

There has been commendable work done in the domain of automated speech scoring systems but due to their black-box nature, many of such systems are rendered useless where the interpretation of the model is important. There are different ways that attempt to interpret machine learning models. The general practice is to use a popular model-agnostic method of interpretability called feature importance. The less popular model-agnostic methods like Partial Dependence Plots and Shapely values can also be of great importance in revealing the impact of input features on the predictions is required to be studied \citep{pdpth,pdpaplication}. Another metric called Shapley \citep{shapth,shape,shap} value which is borrowed from game theory has also proven itself to be a useful method to understand the feature importance, correlation, as well as the feature impact on the model's outcome. Many researchers also perform ablation studies to understand the behavior of the model when the training data is manipulated. To the best of our knowledge, the only type of analysis that has been done on the scoring models is by studying feature importance, Pearson correlations, and ablation studies in which performance of the scoring model is judged after manipulating training data either by adding  \citep{loukina-etal-2017-speech,aspiringminds} or removing \citep{kumar2019get} a feature category from the original set. In this paper, we expand on the use of model-agnostic methods of interpretability and make use of the observations to draw inferences about the linguistic cues that lead to higher proficiency scores. We present feature importance, correlation, and the feature value ranges on the Filipino dataset where the probability of being graded a higher score increases or decreases. Lastly, we confirm that our interpretations of the selected features adhere to the rubrics followed by the human-raters.




\section{Data}
\label{sec:dataset}

For this study, we use the data collected through the administration of Simulated Oral Proficiency Interview (SOPI) \citep{malone2000simulated} for L2 English speakers, primarily from the Philippines. The interview is conducted through an online platform by Second Language Testing, Inc. (SLTI), a US-based language testing company, to screen potential employees. Each candidate is presented with a form containing six prompts of varying difficulty levels. These prompts demand opinions, reasoning, and narration in the form of spontaneous response. For example, one prompt asks the candidate's opinions on decisions made by the government to use forest areas for a new library. The candidate is provided 30 seconds to prepare their response and respond in a stipulated duration (of 60 seconds to 120 seconds depending on the difficulty of prompt) in a sequential manner. These monologue responses are recorded and later scored independently by two human-expert raters. In case of any disagreement between two scores, a third expert is brought in to resolve the conflict. These scores are treated as final. To respond to the prompts, the candidates require explanatory and argumentative abilities. Depending on their performance in providing responses, the candidates are assessed to have the English oral proficiency based on the SLTI Scoring Rubrics (see Section~\ref{subsec:scoring_rubric}). The prompts as well as scoring rubrics are aligned with the guidelines of the Common European Framework of Reference (CEFR) \citep{council2001common}. Many other works have used the SOPI dataset for tasks including automated scoring, sampling, and coherence modeling \citep{grover2020multi,patil2020towards,stansfield2008testing,singla2021speaker,singla2021using}.
Table~\ref{tab:dataset} shares the relevant statistics of the dataset.

\begin{table*}[thb]
  \caption{Statistics of the dataset. P: Prompt Number, \#R: Number of responses, D: Difficulty level, Sz: Average Response Size (Duration in seconds, Length in number of word tokens), and DS: Distribution of Scores (prefixes `L' means Low and `H' means High).}
  \label{tab:dataset}
  \centering
  \small
  \begin{tabular}{|*{10}{c|}}
    \hline 
    \multirow{2}{*}{\textbf{P}} &
    \multirow{2}{*}{\textbf{\#R}} &
    \multirow{2}{*}{\textbf{D}} &
    \multicolumn{2}{c|}{\textbf{Sz}} &
    \multicolumn{5}{c|}{\textbf{DS}} \\
    \cline{4-10}
    &&& \textbf{Duration} & \textbf{Length} & \textbf{A2} & \textbf{LB1} & \textbf{HB1} & \textbf{LB2} & \textbf{HB2} \\
    \hline
    1 & 7877 & B1 & 57.67 & 100.69 & 275 & 1557 & 6045 & - & - \\
    2 & 7432 & B1 & 58.72 & 110.03 & 465 & 2824 & 4143 & - & - \\
    3 & 8042 & B2 & 81.43 & 148.96 & 117 & 664 & 3493 & 3666 & 102 \\
    4 & 8020 & C1 & 104.15 & 180.73 & 121 & 720 & 3536 & 3534 & 109\\
    5 & 7936 & C1 & 105.95 & 196.55 & 110 & 551 & 3004 & 4120 & 151\\
    6 & 8002 & B1 & 55.87 & 109.38 & 119 & 1028 & 6855 & - & -\\
    \hline
  \end{tabular}
\end{table*}2

\subsection{SLTI Scoring Rubrics for SOPI}
\label{subsec:scoring_rubric}
Based on the results of a needs analysis, the minimum level of English proficiency for being able to effectively carry out the work-related tasks was deemed to be the B2 level of the CEFR. This means that SLTI SOPI should be able to reliably identify test takers' performance at B2 level or above as qualified candidates in terms of their speaking ability, and also identify those test takers who could not meet this requirement (due to proficiency lower than B2 in the CEFR). However, rather than merely focusing on a minimum proficiency level, and to make the test scores more informative and broaden their use for a wider range of contexts, the SLTI SOPI includes six tasks to elicit spoken responses that are ratable at the B1 to C1 levels and their respective sub-levels, i.e., Low-B1 (B1.1), High-B1 (B1.2), Low-B2 (B2.1), High-B2 (B2.2), Low-C1 (C1.1), and High-C1 (C1.2). 

Accordingly, the three out of the six SLTI SOPI tasks aim to elicit responses that the CEFR describes as B1. The responses to these B1 tasks are ratable within the range of A2, B1.1, and B1.2 levels. One of the SLTI SOPI tasks was developed with the aim of eliciting responses at the B2-level, and the responses to this task are ratable within the range of A2, B1.1, B1.2, B2.1, and B2.2 levels. Finally, the two remaining SLTI SOPI tasks aim to elicit responses at the C1 level, and the responses to these two tasks are ratable within the range of A2, B1.1, B1.2, B2.1, B2.2, and C1 levels. This way, the responses of the test takers can be rated at each of the levels (except for C1) at least three times, which is the minimum requirement for the reliable assessment of spoken performances at each of the above levels \citep{kenyon2000rating}.


To maximize the representation of the construct of communicative competence, as delineated by \cite{bachman1996language}, and following the underlying construct of ETS Speechrater \citep{xi2008automated}, the holistic rating rubric of the SLTI SOPI includes descriptions of the three main components of delivery, language use, and topic development.

Delivery includes concepts of pace and clarity of the speech and is about whether speech is generally clear, fluid, and sustained. It includes two main features of pronunciation and prosody, which in turn, include sub-features of stress, intonation, pacing and rhythm \citep{hsieh2019features}. In assessing this component of communicative competence, raters consider the test takers' pronunciation, rate of speech, and degree of hesitancy. 

Language Use component includes sub-constructs of diversity, sophistication, and precision of vocabulary use, as well as, the complexity and accuracy of grammar use. For evaluating this aspect of communicative competence, raters evaluate examinees' ability to select words and phrases and their ability to produce structures that appropriately and effectively communicate their ideas.

Finally, Topic Development refers to the coherence and fullness of the response. When assessing this dimension of communicative competence, raters evaluate the progression of ideas and coherence, as well as, the degree of elaboration and completeness of the response. In the SLTI SOPI rating rubric, the description of these aspects of the communicative competence has been mapped on the CEFR level descriptions.

\section{Features And Their Extraction}
In this section, we will discuss our methodology for extracting features from the SOPI dataset in detail. We explain the process of extracting text from the raw audio file and the need for forced alignment of the transcribed text and the audio files in the process of feature extraction. We further go into detail discussing the extraction of the linguistic features across fluency, suprasegmental pronunciation, content, grammar and vocabulary, and acoustic features. Each feature in the said feature category is presented with its origin and definition. Additionally, we provide Pearson correlation scores of the individual features with the human-rated score. We also present the Pearson correlation as reported by ETS on their dataset alongside the correlation on the SOPI dataset for the features we borrowed from their literature.

\label{sec:features}
\subsection{Automatic Speech Recognition}
The non-native audio responses were encoded using an end-to-end DeepSpeech~2 \citep{DBLP:conf/icml/AmodeiABCCCCCCD16} architecture followed by 4-gram language model decoder to generate the transcription. The system is trained on approximately $1,000$ hours of audio sampled from CommonVoice \citep{ardila-EtAl:2020:LREC} and LibriSpeech dataset \citep{7178964} and further fine-tuned on approximately $45$ hours (or 3558 samples) of transcribed non-native spoken responses sampled from our 8-form dataset. These $45$ hours of non-native spoken responses were manually transcribed using audino (\cite{grover2020audino}). The Form 3, used in this study, contributed close to 36\% of the transcribed responses and were not excluded from our study. This system achieved a Word Error Rate (WER) of $20.2\%$ on approximately $10$ hours of unseen spoken responses.

\subsection{Forced Aligner}
Forced aligners use speech audio and text to produce a time-aligned representation of words and phonemes. The purpose of using a forced aligner for this study is to get such time-aligned word level and phoneme level representations which will be utilized in the feature extraction process. Similar to \cite{rhythm}, for getting stress representation of phonemes for the stress-based features we used a forced aligner that can produce stress markings on the vowels (no stress, primary stress, secondary stress, tertiary stress). In this study, we used a pre-trained Montreal Forced Aligner (MFA) \citep{mcauliffe2017montreal} trained on LibriSpeech dataset \citep{7178964} to get time-aligned transcriptions on our dataset.

\subsection{Features}
The features used in this study were extracted using both the speech audio and the transcribed text we got from the trained ASR. We extracted five groups of hand-crafted features to capture language proficiency cues across fluency, suprasegmental pronunciation, content, and, grammar and vocabulary usage. Some of the features were borrowed from the work summarised by ETS in \citep{speechrater} and are cited accordingly. Additionally, we also extracted acoustic features including jitter and shimmer features to study the evolution of spoken information throughout the speech duration. The succeeding paragraphs talk about each feature group in detail.

\subsubsection{Fluency Features (FF)}
One of the major aspects of speech delivery is to examine the construct of fluency of the speech. Fluency can be divided into two types based on the delivery--- breakdown fluency and speed fluency \citep{chflu}. Breakdown fluency deals with different aspects of pauses like number of silences, number of filled pauses per second (disfluency), mean duration of silences in seconds, etc., and speed frequency deal with the speed of speech delivery related features like speaking rate and articulation rate. Speaking rate is calculated across response time which is the total time of the speech when silences from the beginning and end are ignored, while articulation rate uses articulation time which is the summation of time segments where an utterance was spoken. The latter ignores all pauses including the ones that are important to distinguish between the two spoken words. We do not calculate disfluency features like repair fluency and consider it as a feature which will be included in next version of the work. All the features in Table~\ref{Tab:breakdown} and ~\ref{Tab:speedfluency} are borrowed from \cite{chflu}. These tables also include the correlation scores of these features on the SOPI dataset and as reported by ETS on their dataset.

\begin{table}[]
\centering
\caption{List of Breakdown Fluency features. All the features in this table are borrowed from \cite{chflu}) and used by ETS.}
 \label{Tab:breakdown}
\begin{tabular}{|>{\centering\arraybackslash}p{0.28\columnwidth}|p{0.29\columnwidth}|>{\centering\arraybackslash}p{0.13\columnwidth}|>{\centering\arraybackslash}p{0.13\columnwidth}|}
\hline
\textbf{Feature Name}        & \textbf{Definition} & \textbf{Correlation (SOPI)} & \textbf{Correlation (ETS)} \\ \hline
filled\_pause\_rate          & Number of filled pauses (uh, um) per second.                                                                                                          & -0.14                                                                   & -0.23                                                                  \\ \hline
general\_silence            & Number of silences. (silent duration between two words greater than 0.145 seconds)                   & -0.15                                                                   & -0.26                                                                  \\ \hline
mean\_silence                & Mean duration of silences in seconds.                                                                                                                 & -0.28                                                                   & -0.32                                                                  \\ \hline
silence\_absolute\_deviation & Mean absolute difference of silence duration.                                                                                                        & -0.27                                                                   & -0.32                                                                   \\ \hline
SilenceRate1                 & Number of silences divided by total number of words.                                                                                                  & -0.38                                                                   & -0.50                                                                  \\ \hline
SilenceRate2                 & Number of silences divided by total response duration in seconds.                                                                                     & -0.24                                                                   & -0.45                                                                  \\ \hline
long\_silence\_deviation     & Mean deviation of long silences in seconds. (silent duration between two words greater than 0.495 seconds. & -0.16                                                                   & -0.26                                                                  \\
\hline
\end{tabular}
\end{table}

\begin{table}[]
\centering
\caption{List of Speed Fluency features. All the features in this table are borrowed from \cite{chflu} and used by ETS.}
 \label{Tab:speedfluency}
 \begin{tabular}{|>{\centering\arraybackslash}p{0.25\columnwidth}|p{0.30\columnwidth}|>{\centering\arraybackslash}p{0.13\columnwidth}|>{\centering\arraybackslash}p{0.13\columnwidth}|}
\hline
\textbf{Feature Name}        & \textbf{Definition} & \textbf{Correlation (SOPI)} & \textbf{Correlation (ETS)} \\ \hline
speaking\_rate               & Number of words per second in total response duration.                                                                                                & 0.44                                                                    & 0.54                                                                   \\ \hline
articulation\_rate           & Number of words per second in total articulation time.                                                                                                & 0.21                                                                    & 0.38                                                                   \\ \hline
longpfreq        & Frequency of long pauses normalized by
response length in words                                                                                & -0.42                                              & - \\
\hline
\end{tabular}
\end{table}

\begin{table}[]
\caption{List of Stress-based features. All the features in this table are borrowed from \cite{chflu} and used by ETS. }
 \label{Tab:stressbased}
 \begin{tabular}{|>{\centering\arraybackslash}p{0.25\columnwidth}|p{0.30\columnwidth}|>{\centering\arraybackslash}p{0.15\columnwidth}|>{\centering\arraybackslash}p{0.15\columnwidth}|}
\hline
\textbf{Feature Name}  & \textbf{Definition}                                                                                                                   & \textbf{Correlation (SOPI)} & \textbf{Correlation (ETS)} \\ \hline
StressedSyllPercent    & Relative frequency of stressed syllables in percent.                                                                                  & 0.04                                                                    & 0.38                                                                   \\ \hline
StressDistanceSyllMean & Mean distance between stressed syllables in syllables.                                                                                & -0.12                                                                   & -0.37                                                                  \\ \hline
StressDistanceSyllSD  & Mean deviation of distances between stressed syllables in syllables.                        & -0.08                                                                   & -0.33                                                                  \\ \hline
StressDistanceMean     & Mean distance between stressed syllables in seconds.                                                                                  & -0.20                                                                   & -0.47                                                                  \\ \hline
StressDistanceSD       & Mean deviation of distances between stressed syllables in seconds. & -0.16                                                                   & -0.41                                                                  \\ \hline
\end{tabular}
\end{table}

\subsubsection{Suprasegmental Pronunciation Features (SPF)}

The other aspect of speech delivery is studying prosody (stress-based features) and rhythm (interval-based features) within the speech, combined called suprasegmental pronunciation \citep{chflu}. Stress-based features capture occurrences and distances between consecutive stressed phonemes as listed in Table~\ref{Tab:stressbased}. Interval-based features, on the other hand, deal with features that are solely dependent on vocalic, consonantal, and syllabic regions in the speech as listed in Table~\ref{Tab:rhythmbased}. Most of the features listed Table~\ref{Tab:stressbased} and \ref{Tab:rhythmbased} are calculated using literature from \cite{chflu} except the ones that are marked as * which are introduced by us. We also mention the correlation scores for the borrowed feature on both the SOPI dataset and the ETS dataset in the given tables.

Please note that the correlation for stress-based features is lower for SOPI data when compared against ETS data. This can be owned to the difference in the process followed to calculate these values. While ETS trained their own forced aligner and used a decision tree to capture both the acoustic and linguistic information, we were constrained to use MFA which is dependent on static stress based information.

\begin{table}[]
\centering
\caption{List of Interval-based features. (* indicates newly introduced features while the rest are borrowed from \cite{chflu} and used by ETS)}
 \label{Tab:rhythmbased}
 \begin{tabular}{|>{\centering\arraybackslash}p{0.30\columnwidth}|p{0.25\columnwidth}|>{\centering\arraybackslash}p{0.15\columnwidth}|>{\centering\arraybackslash}p{0.15\columnwidth}|}
\hline
\textbf{Feature Name}  & \textbf{Definition}                                                                                                                   & \textbf{Correlation (SOPI)} & \textbf{Correlation (ETS)} \\ \hline
vowelPercentage        & Percentage of speech that consists of vowels.                                                                                         & -0.24                                                                   & -0.30                                                                  \\ \hline
vowelDurationSD        & Standard Deviation of vowel segments.                                                                        & -0.15                                                                   & -0.26                                                                  \\ \hline
consonantDurationSD    & Standard Deviation of consonantal segments.                                                                                           & -0.09                                                                   & -0.20                                                                  \\ \hline
syllableSDNorm         & Standard Deviation of syllable segments divided by mean length of syllable segments.       & -0.15                                                                   & -0.24                                                                  \\ \hline
vowelPVI               & Raw Pairwise Variability Index for vocalic segments.                                                                                  & -0.19                                                                   & -0.39                                                                  \\ \hline
consonantPVI           & Raw Pairwise Variability Index for consonantic segments.                                                                              & -0.13                                                                   & -0.36                                                                  \\ \hline
syllablePVI            & Raw Pairwise Variability Index for syllable segments.                                                                                 & -0.19                                                                   & -0.4                                                                   \\ \hline
vowelPVINorm           & Normalized Pairwise Variability Index for vocalic segments.                                                                           & -0.19                                                                   & -0.25                                                                  \\ \hline
consonantPVINorm       & Normalized Pairwise Variability Index for consonantic  segments.                            & -0.24                                                                   & -0.32                                                                  \\ \hline
syllablePVINorm        & Normalized Pairwise Variability Index for syllable segments.                                                                          & -0.25                                                                   & -0.29                                                                  \\ \hline
vowelSDNorm*            & Standard Deviation of vowel segments divided by mean length of vowel segments.            & -0.17                                                                   & -                                                                      \\ \hline
syllableDurationSD*     & Standard Deviation of syllable segments.                                                                                              & -0.14                                                                   & -                                                                      \\ \hline
consonantSDNorm*        & Standard Deviation of consonantal segments divided by mean length of consonantic segments. & -0.13                                                                   & -                                                                      \\ \hline
consonantPercentage*    & Percentage of speech that consists of consonants.                                                                                     & 0.40                                                                    & -                                                                      \\ \hline
\end{tabular}
\end{table}

\subsubsection{Content Features (CF)}

The generally accepted way of capturing content information in scoring work has been to find overlap between the reference content and the test taker's response \citep{chcontent}. While the reference content in most of the cases is the question prompt itself, the same was not feasible in our work due to the shorter question length and the open-ended nature of the asked questions. Thus, instead of finding content overlap (reference-based features), we considered response-based features extracting term frequencies weighted by inverse document frequency values (TF-IDF) word vectors for each response. These word vectors were extracted on the transcribed text we got from the trained ASR.

\subsubsection{Grammar  and Vocabulary Features (GVF)}
Grammar and vocabulary usage marks the next major component of language proficiency. It helps assess the responses across the dimensions of accuracy and sophistication of language construct. We make use of L2 Syntactic Complexity Analyzer \citep{lu,syntactic} and Lexical Complexity Analyser \citep{tool} to extract features like lexical diversity, lexical sophistication, verb sophistication, type to token ratio, dependent clause per clause, coordinate phrase per clause etc. as listed in Table~\ref{Tab:lexicalcomplexity} and \ref{Tab:syntacticcomplexity}. Please note that these tools require a punctuated texts to produce the required values. To achieve this we used an open-source punctuator tool, called punctuator2 \citep{tilk2016}, which added the required punctuation to the text before being sent to the complexity analyzer tools. \cite{wordlist} presents a word complexity lexicon list where each word is given a complexity score rating by eleven annotators. Each complexity score is between the scale of one to six, where one represents ``very simple" and six represents ``very complex". To make use of this list we calculated average and mode values of the scores annotated by the human-raters for each lexical. We do so to create variants of text complexity based features, one that uses average values, and the one that makes use of mode values of the complexity scores. Using thus formed word list, we extract features like average complexity of spoken content when stop words are removed and when stop words are kept intact. Additionally, we also take into account the count-based features (as listed in Table~\ref{Tab:countbased}) that deal with the total frequency of POS tag occurrences in the response like total adverbs, total nouns, total conjunctions, etc. 

\begin{table}[]
\centering
\caption{List of Lexical Complexity features extracted using Lexical Analyser Tool \citep{tool}.}
 \label{Tab:lexicalcomplexity}
 \begin{tabular}{|>{\centering\arraybackslash}p{0.2\columnwidth}|p{0.35\columnwidth}|>{\centering\arraybackslash}p{0.15\columnwidth}|}
\hline
\textbf{Feature Name}  & \textbf{Definition} & \textbf{Correlation (SOPI)} \\ \hline
ld                                      & Lexical diversity.                                                  & 0.01                                                                    \\ \hline
ls1                                     & Lexical Sophistication -I                                           & 0.04                                                                    \\ \hline
ls2                                     & Lexical Sophistication -II                                          & 0.13                                                                    \\ \hline
vs1                                     & Verb Sophistication-I                                               & 0.04                                                                    \\ \hline
vs2                                     & Verb Sophistication-II                                              & 0.04                                                                    \\ \hline
ndw                                     & Number of different words.                                          & 0.44                                                                    \\ \hline
ndwz                                    & Number of different words from first 50 words.                      & 0.31                                                                    \\ \hline
ndwerz                                  & Number of different words from expected random 50 words.            & 0.37                                                                    \\ \hline
ndwesz                                  & Number of different words from expected sequence of 50 words.       & 0.35                                                                    \\ \hline
ttr                                     & Type to Token Ratio.                                                & 0.16                                                                    \\ \hline
\end{tabular}
\end{table}

\begin{table}[]
\centering
\caption{List of Syntactic Complexity features extracted using Syntactic Analyser Tool \citep{lu}.}
 \label{Tab:syntacticcomplexity}
 \begin{tabular}{|>{\centering\arraybackslash}p{0.2\columnwidth}|p{0.35\columnwidth}|>{\centering\arraybackslash}p{0.15\columnwidth}|>{\centering\arraybackslash}p{0.15\columnwidth}|}
\hline
\textbf{Feature Name}  & \textbf{Definition} & \textbf{Correlation (SOPI)} \\ \hline
MLS                                     & Mean length of sentence.                                            & 0.07                                                                    \\ \hline
MLT                                     & Mean length of T-unit.                                              & 0.02                                                                    \\ \hline
MLC                                     & Mean length of clause.                                              & -0.04                                                                   \\ \hline
C/T                & Number of clauses per sentence.                                     & 0.06                                                                    \\ \hline
VP/T                                    & Number of verb phrases per T-unit.                                  & 0.01                                                                    \\ \hline
DC/C                                    & Number of dependent clauses per clauses.                            & 0.1                                                                     \\ \hline
DC/T                                    & Number of dependent clauses per T-unit.                             & -0.03                                                                   \\ \hline
T/S                                     & T-units per sentence.                                               & 0.05                                                                    \\ \hline
CT/T                                    & Complex T-unit per T-unit.                                          & 0.06                                                                    \\ \hline
CP/T                                    & Coordinate Phrase per T-unit.                                       & -0.03                                                                   \\ \hline
CP/C                                    & Coordinate Phrase per clause.                                       & -0.05                                                                   \\ \hline
CN/T                                    & Complex Nominal per T-unit.                                         & 0.01                                                                    \\ \hline
CN/C                                    & Complex Nominal per clause.                                         & -0.03                                                                   \\ \hline
\end{tabular}
\end{table}

\begin{table}[]
\centering
\caption{List of Count-based and text-complexity features. Other than * which indicates newly introduced features, all features are extracted using Syntactic Analyser Tool \citep{lu}.}
 \label{Tab:countbased}
 \begin{tabular}{|>{\centering\arraybackslash}p{0.45\columnwidth}|p{0.25\columnwidth}|>{\centering\arraybackslash}p{0.15\columnwidth}|}
\hline
\textbf{Feature Name}  & \textbf{Definition} & \textbf{Correlation (SOPI)} \\ \hline

total\_adjectives*                       & Total number of adjectives.                                          & 0.3                                                                     \\ \hline
total\_adverbs*                          & Total number of adverbs.                                            & 0.4                                                                     \\ \hline
total\_nouns*                            & Total number of nouns.                                              & 0.31                                                                    \\ \hline
total\_verbs*       & Total number of verbs.                                              & 0.36                                                                    \\ \hline
total\_pronoun*                          & Total number of pronouns.                                           & 0.28                                                                    \\ \hline
total\_conjunctions*                     & Total numver of conjunctions.                                       & 0.37                                                                    \\ \hline
total\_determiners*                      & Total number of duration.                                           & 0.29                                                                    \\ \hline
total\_text\_complexity\_no\_sw\_mAvg*   & Total text complexity (average values) when stop words are removed.                  & 0.39                                                                    \\ \hline
average\_word\_complexity\_no\_sw\_mAvg* & Average text complexity (average values) when stop words are removed.                & 0.03                                                                    \\ \hline
total\_text\_complexity\_mAvg*           & Total text complexity (average values) when no stop words are removed.               & 0.39                                                                    \\ \hline
average\_word\_complexity\_mAvg*         & Average text complexity (average values) when no stop words are removed.             & 0.04                                                                    \\ \hline
average\_syllables\_in\_words*           & Average syllables in a word excluding (average values) stop words from the response. & 0.06                                                                    \\ \hline
total\_text\_complexity\_no\_sw\_mMod*   & Total text complexity (mod values) when stop words are removed.                  & 0.39                                                                    \\ \hline
average\_word\_complexity\_no\_sw\_mMod* & Average text complexity (mod values) when stop words are removed.                & 0.03                                                                    \\ \hline
total\_text\_complexity\_mMod*           & Total text complexity (mod values) when no stop words are removed.               & 0.39                                                                    \\ \hline
average\_word\_complexity\_mMod*         & Average text complexity (mod values) when no stop words are removed.             & 0.04                                                                    \\ \hline
average\_syllables\_in\_words*           & Average syllables in a word excluding stop words from the response. & 0.06                                                                    \\ \hline
W                                       & Total number of words.                                              & 0.38                                                                    \\ \hline
VP                                      & Total number of Verb Phrases.                                       & 0.34                                                                    \\ \hline
C                                       & Total number of Clauses.                                            & 0.35                                                                    \\ \hline
T                                       & Total number of T-units.                                            & 0.3                                                                     \\ \hline
DC                                      & Total number of Dependent Clauses.                                  & 0.32                                                                    \\ \hline
CT                                      & Total number of Complex T-units.                                    & 0.32                                                                    \\ \hline
CP                                      & Tota number of Coordinate Phrases.                                  & 0.2                                                                     \\ \hline
CN                                      & Total number of Complex Nominals.                                   & 0.34                                                                    \\ \hline

\end{tabular}
\end{table}

\subsubsection{Acoustic Features (AF)}
So far all the features we extracted were time-aggregated. As mentioned in \cite{7404814}, time-aggregated features do not fully consider the evolution of the response over time. Thus, we extracted time-sequenced prosodic features like variants of jitter and shimmer, that can capture the evolution of information throughout the speech duration. We used Librosa \citep{librosa} and Praat \citep{boersma2001speak} scripts to extract features like loudness, jitter, shimmer, and also pitch and energy-based acoustic features. We also included the total duration of the speech as a feature in this feature group. Table~\ref{tab:acoustic_features} lists the features, their definitions and correlation.

\begin{table}[htb]
\centering
\caption{List of acoustic features. * indicates features borrowed from ETS SpeechRater\textit{\textsuperscript{SM}} \citep{speechrater}.}
\label{tab:acoustic_features}
\begin{tabular}{|l|l|c|}
\hline
\textbf{Feature Name}                                              
& \textbf{Definition}                                                                                                                                                           
& \textbf{Correlation (SOPI)} \\ 
\hline
stdev\_energy                       & \begin{tabular}[c]{@{}l@{}}Standard deviation of energy of \\ the response.\end{tabular}  & 0.04  \\ \hline
mean\_pitch                         & Mean pitch of the response.                                                               & -0.05 \\ \hline
stdev\_pitch                        & \begin{tabular}[c]{@{}l@{}}Standard deviation of pitch of \\ the response.\end{tabular}   & -0.04 \\ \hline
range\_pitch*                      & Range of pitch of the response.                                                           & 0.14  \\ \hline
zero\_crossing\_rate                & \begin{tabular}[c]{@{}l@{}}Rate of sign change across the \\ audio signal.\end{tabular}    & -0.05 \\ \hline
energy\_entropy                     & \begin{tabular}[c]{@{}l@{}}Entropy of the normalised energy\\ of sub-frames of the audio signal.\end{tabular}     & 0.24 \\\hline
spectral\_centroid                  & \begin{tabular}[c]{@{}l@{}}Weighted average of all the \\ frequencies in the given \\ response signal. It is closely \\ related to the brightness of a\\  sound.\end{tabular} & -0.01                                                                 \\ \hline
\begin{tabular}[c]{@{}l@{}}Jitter and \\ its variants\end{tabular}  & \begin{tabular}[c]{@{}l@{}}Measure of frequency instability. \\ Variants that were calculated are\\  rapJitter, ppq5Jitter, and ddpJitter.\end{tabular}                       & -0.03                                                                 \\ \hline
\begin{tabular}[c]{@{}l@{}}Shimmer and \\ its variants\end{tabular} & \begin{tabular}[c]{@{}l@{}}Measure of amplitude instability. \\ Variants that were calculated are \\ localShimmer, apq3Shimmer, \\ aqpq5Shimmer and ddaShimmer.\end{tabular}  & -0.04                                                                 \\ \hline
\end{tabular}
\end{table}

All the features we discussed above are hand-crafted such that they capture meaningful cues of language proficiency. We develop a pipeline to extract all of these features for all the prompts in the SOPI dataset. The reason for limiting the feature space only to the linguistic features for this task is to use the model for interpretation on these features and observe the effect on proficiency scores when the feature values are altered.

\section{Experimentations}
\label{sec:experimental_setup}

\cite{classreg} discussed the advantages and disadvantages of using classification and regression approaches for formulating the essay scoring problem. The multi-class approach of classification does not consider the order of the classes during the training process. This becomes a disadvantage as the order of classes is an intrinsic property of speech scoring task. The regression approach takes the order of a class into consideration but the distances between adjacent classes may not always be the same. This calls for a careful transformation of the numeric labels to the values for regression. Through results, \cite{classreg} show that regression outperforms the classification formulation of the essay scoring. Inspired from \cite{classreg}, we also perform experiments for both classification and regression analysis to observe which formulation of speech scoring tasks on the SOPI dataset works more efficiently. We train multiple machine learning models including Decisions Tree, Random Forest, Support Vector Machine, Gradient Boosted Trees, and XGBoost for both classification and regression tasks. Additionally, we train Linear Regression and Logistic Regression as baselines for the regression and classification tasks respectively. We also train Random Forest baseline with only length of response as predictor to check dependence of grades on number of words being spoken. We evaluate the performance of our models based on the evaluation metrics mentioned in Section~\ref{sec:evaluation_metric}. We make use of the same metrics to select the best formulation of speech scoring specific to our dataset. The model thus selected will be utilized for feature analysis and model interpretability tasks as discussed in Section 6.

\subsection{Setup}

\begin{table*}[thb]
  \caption{Statistics of the splits created. P: Prompt Number, ST: Split type, \#R: Number of Responses, MS: Mean Score and DS: Distribution of Scores (prefixes `L' means Low and `H' means High; Regression mapped scores mentioned in brackets).}
  \label{tab:splits}
  \centering
  \small
  \begin{tabular}{|*{9}{c|}}
    \hline 
    \multirow{2}{*}{\textbf{P}} &
    \multirow{2}{*}{\textbf{ST}} &
    \multirow{2}{*}{\textbf{\#R}} &
    \multirow{2}{*}{\textbf{MS}} &
    \multicolumn{5}{c|}{\textbf{DS}} \\
    \cline{5-9}
    &&&& \textbf{A2 (0)} & \textbf{LB1 (1)} & \textbf{HB1 (2)} & \textbf{LB2 (3)} & \textbf{HB2 (4)} \\
    \hline
    \multirow{3}{*}{1} & train & 5670 & 1.732 & 185 & 1152 & 4333 & - & - \\
                       & valid & 631 & 1.737 & 27 & 112 & 492 & - & -  \\
                       & test & 1576 & 1.734 & 63 & 293 & 1220 & - & - \\
    \hline
    \multirow{3}{*}{2} & train & 5176 & 1.496 & 319 & 1973 & 2884 & - & -  \\
                       & valid & 808 & 1.499 & 63 & 279 & 466 & - & -  \\
                       & test & 1448 & 1.49 & 83 & 572 & 793 & - & -  \\
    \hline
    \multirow{3}{*}{3} & train & 5641 & 2.375 & 75 & 472 & 2423 & 2602 & 69  \\
                       & valid & 821 & 2.378 & 13 & 66 & 353 & 376 & 13  \\
                       & test & 1580 & 2.344 & 29 & 126 & 717 & 688 & 20  \\
    \hline
    \multirow{3}{*}{4} & train & 5774 & 2.341 & 93 & 529 & 2549 & 2521 & 82  \\
                       & valid & 642 & 2.364 & 7 & 58 & 284 & 280 & 13  \\
                       & test & 1604 & 2.365 & 21 & 133 & 703 & 733 & 14  \\
    \hline
    \multirow{3}{*}{5} & train & 5713 & 2.451 & 89 & 406 & 2159 & 2957 & 102  \\
                       & valid & 635 & 2.465 & 6 & 37 & 265 & 310 & 17  \\
                       & test & 1588 & 2.491 & 15 & 108 & 580 & 853 & 32  \\
    \hline
    \multirow{3}{*}{6} & train & 5760 & 1.839 & 92 & 744 & 4924 & - & -  \\
                       & valid & 641 & 1.861 & 5 & 79 & 557 & - & -  \\
                       & test & 1601 & 1.844 & 22 & 205 & 1374 & - & -  \\
    \hline
  \end{tabular}
\end{table*}

To prepare the training dataset, we first standardize dataset features for all the responses of each six prompts. Then, we create a stratified train, validation, and test split with a ratio of $70:10:20$. Table~\ref{tab:splits} shares the statistics of the splits created. Using this data, we train all the models mentioned above. To compare the performance of the models the choice of evaluation metrics remains the weighted Quadratic Kappa Score (QWK), Pearson's R correlation (\textit{r}), and the Mean Squared Error (MSE). These evaluation metrics are calculated and reported on the test set of prompts for every model. We perform extensive hyper-parameter tuning on each model using the Grid Search algorithm with 5-fold cross-validation. Based on the close analysis of the results from the evaluation metrics, we first choose the ideal formulation of the speech scoring task followed by selecting the best non-neural model for performing further experiments. Also, please note that the data in itself is highly imbalanced with very low data points for lower proficiency levels, thus the model may favor the dominant class. To reduce such a biased influence on the model, class weights were introduced during the training phase wherever necessary.



\subsection{Evaluation Metrics}
\label{sec:evaluation_metric}

To evaluate the models trained, we use the Quadratic Weighted Kappa (QWK) score as the primary metric. Previous work have used QWK score as primary metric for automated essay \citep{taghipour-ng-2016-neural,AAAI1816431} and short answer \citep{riordan-etal-2017-investigating,kumar2019get} scoring tasks. The QWK score measures the inter-rater agreement, which for our task measures agreement between the predicted grade by the trained models and final human expert grade. It normally ranges from $0$ to $1$ and can also be negative if there is lesser agreement than expected by chance. We calculate the QWK score by constructing a weighted matrix $W$ of size $N~\times~N$, where $N$ represents the number of classes, using the formula as mentioned in Equation~\ref{Eq:weight_matrix}.

\begin{equation}
\label{Eq:weight_matrix}
  W_{i,j} = \frac{(i-j)^{2}}{(N-1)^{2}}
\end{equation}

Here, $i$ and $j$ represent the human expert grade and the predicted grade by model, respectively. We, then, construct a confusion matrix $O$ of size $N~\times~N$, where $O_{ij}$ equals to the number of speech responses that receive grade $i$ by the human and grade $j$ by the model.

We construct a histogram matrix of expected grades $E$. This is computed as the outer product between the histogram vector of actual grades and the histogram vector of predicted grades. This is followed by normalization to ensure $E$ and $O$ have the same sum.

Lastly, the QWK ($\kappa$) is obtained as in Equation~\ref{Eq:qwk}.
\begin{equation}
\label{Eq:qwk}
  \kappa = 1 - \frac{\sum_{i,j}W_{i,j}O_{i,j}}{\sum_{i,j}W_{i,j}E_{i,j}}
\end{equation}

We also track the Pearson correlation coefficient \citep{aspiringminds,speechrater} and Mean Squared Error (MSE) as secondary metrics. Pearson correlation measures the linear relationship between predicted grades and corresponding actual grades and varies between $-1$ to $+1$ with $0$ implying no correlation and extremes implying perfect linear relationship. Positive correlation coefficient means as predicted grades increase in value, the human grades increase, and likewise negative correlation coefficient means as predicted grades increase, the human grades decrease. Mean Squared Error (MSE) helps measure the quality of the estimator. Here, the closer the MSE is to $0$, the better is the estimator. It is calculated as shown in Equation~\ref{Eq:mse}.

\begin{equation}
\label{Eq:mse}
  MSE = \frac{1}{N}\sum_{i=1}^{N}(y_{true}^{i} - y_{pred}^{i})^{2}
\end{equation}

Here, $N$ is the number of responses in the evaluation set, $y_{true}^{i}$ and $y_{pred}^{i}$ are the human and predicted grades, respectively, for the $i^{th}$ response.

\section{Results and Analysis}
\label{sec:results}

We report QWK, \textit{r} and MSE for each of the model trained for classification and regression analysis of the speech scoring task in  Table~\ref{Tab:regression_results} and \ref{Tab:classification_results}. We keep Logistic Regression and Linear Regression as the baselines for classification and regression analysis, respectively. On comparison, we notice that for regression analysis, XGBoost outperforms every other model by achieving an average QWK of $0.475$, \textit{r} of $0.493$, and MSE of $0.299$ across all the six prompts. On the other hand, for classification analyses, XGBoost seems to be performing with the highest scores of QWK and MSE for all the prompts except for Prompt 1 and 2 where the Gradient Boosted Trees seems to perform better. For classification analysis, we pick the XGBoost model as the best performing model with a higher average QWK of $0.458$, \textit{r} of $0.484$, and MSE of $0.333$. For the overall best performing model for the speech scoring task, we choose the final formulation on which further analysis and interpretation will be performed. Based on the average values of reported evaluation metrics, we observe that regression analysis performs better for XGBoost when compared with classification analysis. \cite{classreg} has made similar observations for essay scoring tasks, sharing a more thorough evaluation. Since this comparison between classification and regression analysis is an observation, specific to our dataset, for now, we limit ourselves to experiment with both the approaches to compare the results and pick the final formulation for speech scoring on our dataset. Thus, we pick the XGBoost regressor as the final model for the speech scoring task and further analysis.

We observe that Prompt 2 performance for all the models including XGBoost regressor has been on the lower end when compared with the other prompts. The probable reason for that can be attributed to the poor inter-annotator QWK agreement of $0.560$ compared to the average QWK of $0.758$ across the other prompts. We also observe the Random Forest baseline, which uses only number of words as predictor, performs poorly compared to other models, justifying the dependence of scores on other features introduced previously.

The prediction performance of speech assessment systems is lower when compared to written assessment systems. To our knowledge, there has been no formal studies made which compare the two systems. \cite{speechrater}, in their introduction, talk about the impact of speaking characteristics like native language and proficiency level on the performance of automatic speech recognition systems. This influence of speech recognition accuracy on automatic speech scoring systems was studied in detail by \cite{7078590}, where they vary WER of ASR transcription and analyze the drop in correlation between human and automated scores. Environmental factors such as noise and multi-speaker speech also impacts the performance of speech recognition and remains a challenge \citep{10.1145/3178115}. Limited public availability of L2 English spontaneous data has also been a key challenge in degraded performance of ASR systems. Finally, written text is much cleaner and is editable, something that is not possible in case of speech responses, which contains disfluencies.

\begin{landscape}
    \begin{table}[h]
        \centering
        \setlength{\tabcolsep}{2pt}
        \caption{Quadratic Kappa (QWK) Score, Pearson correlation (\textit{r}) and Mean Squared Error (MSE) across prompts for regression models. Gradient Boosted Trees (GBT), Random Forest (RF), Support Vector Machine (SVM), Linear Regression (LR), Random Forest with only length of response as predictor (Baseline) and Human-Human (HH) results are presented.}
        \label{Tab:regression_results}
        \begin{tabular}{|*{19}{c|}}
        
        \hline
        \multirow{2}{*}{\textbf{Model}} &
        \multicolumn{3}{c}{\textbf{Prompt 1}}    & 
        \multicolumn{3}{c}{\textbf{Prompt 2}}    & 
        \multicolumn{3}{c}{\textbf{Prompt 3}}    & 
        \multicolumn{3}{c}{\textbf{Prompt 4}}    & 
        \multicolumn{3}{c}{\textbf{Prompt 5}}    & 
        \multicolumn{3}{c|}{\textbf{Prompt 6}}    \\
        
        \cline{2-19}
        & \textbf{QWK} & \textbf{\textit{r}} & \textbf{MSE} 
        & \textbf{QWK} & \textbf{\textit{r}} & \textbf{MSE} 
        & \textbf{QWK} & \textbf{\textit{r}} & \textbf{MSE} 
        & \textbf{QWK} & \textbf{\textit{r}} & \textbf{MSE} 
        & \textbf{QWK} & \textbf{\textit{r}} & \textbf{MSE} 
        & \textbf{QWK} & \textbf{\textit{r}} & \textbf{MSE} \\
        \hline
        \textbf{XGBoost}                                                    & \textbf{0.520}        & \textbf{0.550}      & \textbf{0.211}        & \textbf{0.298}        & \textbf{0.306}      & \textbf{0.434}        & \textbf{0.498}        & \textbf{0.512}      & \textbf{0.430}        & \textbf{0.557}        & \textbf{0.568}      & \textbf{0.371}        & \textbf{0.536}        & \textbf{0.549}      & \textbf{0.371}        & \textbf{0.443}        & \textbf{0.478}      & \textbf{0.136}       \\
        
        \textbf{GBT}                                                            
        & 0.495        & 0.529      & 0.220        
        & 0.298        & 0.307      & 0.434        
        & 0.290        & 0.307      & 0.434           
        & 0.495        & 0.510      & 0.432        
        & 0.535        & 0.547      & 0.372        
        & 0.441        & 0.482      & 0.135        \\
        
        \textbf{RF}                                                  
        & 0.517        & 0.554      & 0.210        
        & 0.289        & 0.301      & 0.441        
        & 0.479        & 0.495      & 0.447        
        & 0.550        & 0.563      & 0.370        
        & 0.530        & 0.545      & 0.374        
        & 0.396        & 0.442      & 0.142        \\
        
        \textbf{SVM}                                                      
        & 0.419        & 0.424      & 0.287        
        & 0.148        & 0.156      & 0.521        
        & 0.237        & 0.344      & 0.544
        & 0.303        & 0.384      & 0.496        
        & 0.347        & 0.406      & 0.508        
        & 0.420        & 0.422      & 0.181   \\
        
        \textbf{LR}                                                      
        & 0.303        & 0.298      & 0.433        
        & 0.130        & 0.092      & 0.623        
        & 0.344        & 0.321      & 0.813        
        & 0.395        & 0.310      & 0.744        
        & 0.404        & 0.396      & 0.705      
        & 0.284        & 0.163      & 0.316  \\
        \textbf{Baseline}                                                      
        & 0.368        & 0.412      & 0.268        
        & 0.149        & 0.152      & 0.532        
        & 0.277        & 0.287      & 0.592        
        & 0.288        & 0.301      & 0.550        
        & 0.293        & 0.300      & 0.558      
        & 0.217        & 0.275      & 0.170  \\
        
        \hline
        \textbf{HH}                                                      
        & 0.685        & 0.685      & 0.156        
        & 0.560        & 0.561      & 0.319        
        & 0.781        & 0.781      & 0.216        
        & 0.808        & 0.808      & 0.195        
        & 0.836        & 0.837      & 0.16      
        & 0.683        & 0.683      & 0.094  \\
        \hline
        \end{tabular}
        
        \bigskip
        \caption{Quadratic Kappa (QWK) Score, Pearson correlation (\textit{r}) and Mean Squared Error (MSE) across prompts for classification models. Gradient Boosted Trees (GBT), Random Forest (RF), Support Vector Machine (SVM), Logistic Regression (LR), Random Forest with only length of response as predictor (Baseline) and Human-Human (HH) results are presented.}
       
        \label{Tab:classification_results}
        
        \begin{tabular}{|*{19}{c|}}
        \hline
        \multirow{2}{*}{\textbf{Model}} &
        \multicolumn{3}{c}{\textbf{Prompt 1}}    
        & \multicolumn{3}{c}{\textbf{Prompt 2}}    
        & \multicolumn{3}{c}{\textbf{Prompt 3}}    
        & \multicolumn{3}{c}{\textbf{Prompt 4}}    
        & \multicolumn{3}{c}{\textbf{Prompt 5}}    
        & \multicolumn{3}{c|}{\textbf{Prompt 6}}    \\
        
        \cline{2-19}
         & \textbf{QWK} &\textbf{\textit{r}}  & \textbf{MSE} 
         & \textbf{QWK} & \textbf{\textit{r}} & \textbf{MSE} 
         & \textbf{QWK} & \textbf{\textit{r}} & \textbf{MSE} 
         & \textbf{QWK} & \textbf{\textit{r}} & \textbf{MSE} 
         & \textbf{QWK} & \textbf{\textit{r}} & \textbf{MSE} 
         & \textbf{QWK} & \textbf{\textit{r}} & \textbf{MSE} \\
        \hline
        
        \textbf{XGBoost}                                                        
        & 0.473               & 0.512               & 0.230
        & 0.254               & 0.278               & 0.472                 & 
        \textbf{0.509}        & \textbf{0.527}      & \textbf{0.432}        & \textbf{0.556}        & \textbf{0.558}      & \textbf{0.377}        & \textbf{0.529}        & \textbf{0.547}      & \textbf{0.374}        & \textbf{0.427}        & \textbf{0.487}      & \textbf{0.133}        \\
        
        \textbf{GBT}                                                        
        & \textbf{0.481}        & \textbf{0.539}      & \textbf{0.218}        
        & \textbf{0.260}        & \textbf{0.293}      & \textbf{0.467}       
        & 0.501                 & 0.520               & 0.438        
        & 0.528                 & 0.542               & 0.393       
        & 0.526                 & 0.529               & 0.401        
        & 0.424                 & 0.477               & 0.136        \\
        
        \textbf{RF}                                                  
        & 0.401        & 0.481      & 0.244        
        & 0.235        & 0.282      & 0.474        
        & 0.452        & 0.473      & 0.472        
        & 0.488        & 0.506      & 0.418        
        & 0.456        & 0.482      & 0.416        
        & 0.456        & 0.482      & 0.416        \\
        
        \textbf{SVM}                                                      
        & 0.512        & 0.514      & 0.245        
        & 0.246        & 0.262      & 0.481        
        & 0.497        & 0.502      & 0.475        
        & 0.549        & 0.553      & 0.406        
        & 0.523        & 0.527      & 0.409        
        & 0.481        & 0.481      & 0.160       \\
        
        \textbf{LR}                                                      
        & 0.310        & 0.314      & 0.427        
        & 0.140        & 0.142      & 0.680        
        & 0.375        & 0.376      & 0.713        
        & 0.355        & 0.360      & 0.750        
        & 0.348        & 0.350      & 0.694      
        & 0.263        & 0.269      & 0.280 \\
        
        \textbf{Baseline}                                                      
        & 0.201        & 0.232      & 0.710        
        & 0.098        & 0.104      & 0.916        
        & 0.182        & 0.208      & 1.644        
        & 0.226        & 0.266      & 1.603       
        & 0.170        & 0.206      & 1.799      
        & 0.117        & 0.162      & 0.696 \\
        
        \hline
        \textbf{HH}                                                      
        & 0.685        & 0.685      & 0.156        
        & 0.560        & 0.561      & 0.319        
        & 0.781        & 0.781      & 0.216        
        & 0.808        & 0.808      & 0.195        
        & 0.836        & 0.837      & 0.160      
        & 0.683        & 0.683      & 0.094  \\
        \hline
        \end{tabular}
    \end{table}
    \vspace*{\fill}
\end{landscape}

\subsection{Ablation Study and Feature Importance}
\label{sec:ablation_featureimportance}
\begin{figure}[htp]
\centering

\begin{subfigure}{0.8\textwidth}
    \centering
    \includegraphics[width=\linewidth]{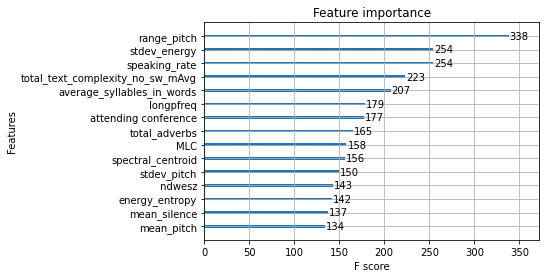}
\end{subfigure}%
\caption{Feature importance of XGBoost Regressor for Prompt 1.}
\label{fig: Feature importance Prompt 1}
\end{figure}

\begin{figure}[htp]
\centering

\begin{subfigure}{0.8\textwidth}
    \centering
    \includegraphics[width=\linewidth]{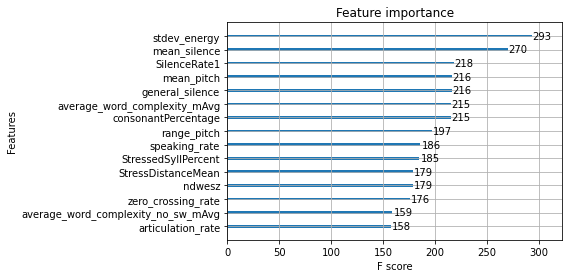}
\end{subfigure}%
\caption{Feature importance of XGBoost Regressor for Prompt 2.}
\label{fig: Feature importance Prompt 2}
\end{figure}

\begin{figure}[htp]
\centering

\begin{subfigure}{0.8\textwidth}
    \centering
    \includegraphics[width=\linewidth]{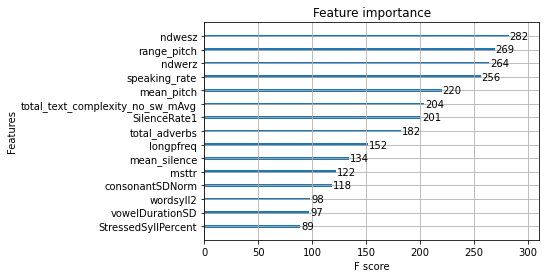}
\end{subfigure}%
\caption{Feature importance of XGBoost Regressor for Prompt 3.}
\label{fig: Feature importance Prompt 3}
\end{figure}

\begin{figure}[htp]
\centering

\begin{subfigure}{0.8\textwidth}
    \centering
    \includegraphics[width=\linewidth]{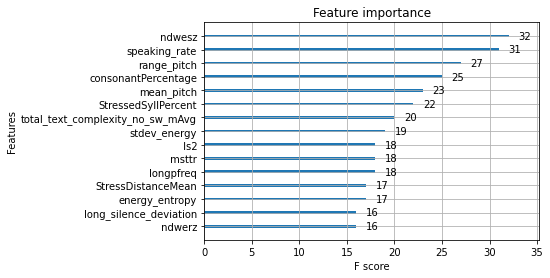}
\end{subfigure}%
\caption{Feature importance of XGBoost Regressor for Prompt 4.}
\label{fig: Feature importance Prompt 4}
\end{figure}

\begin{figure}[htp]
\centering

\begin{subfigure}{.8\textwidth}
    \centering
    \includegraphics[width=\linewidth]{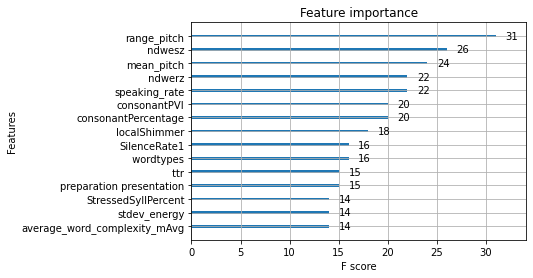}
\end{subfigure}%
\caption{Feature importance of XGBoost Regressor for Prompt 5.}
\label{fig: Feature importance Prompt 5}
\end{figure}

\begin{figure}[htp]
\centering

\begin{subfigure}{0.8\textwidth}
    \centering
    \includegraphics[width=\linewidth]{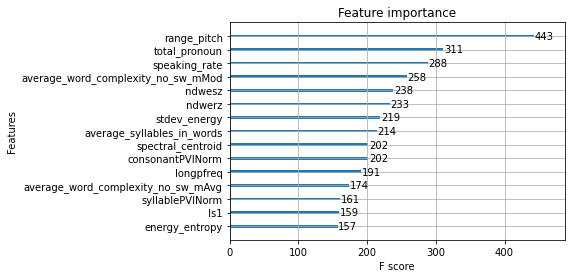}
\end{subfigure}%
\caption{Feature importance of XGBoost Regressor for Prompt 6.}
\label{fig: Feature importance Prompt 6}
\end{figure}

During the training phase of all the models mentioned in Section~\ref{sec:experimental_setup}, we utilized all feature categories for the training purposes. To know the impact of each feature in the model's decision making one of the common approaches is to find the feature importance which is given by the trained model. We present feature importance plots for the XGBoost regressor for all the six prompts in Figures~\ref{fig: Feature importance Prompt 1}, ~\ref{fig: Feature importance Prompt 2},  ~\ref{fig: Feature importance Prompt 3}, ~\ref{fig: Feature importance Prompt 4}, ~\ref{fig: Feature importance Prompt 5} and ~\ref{fig: Feature importance Prompt 6}, when all feature categories are used together. Though this method is successful to get feature importance of individual features, it does not account for the importance of a complete feature set for a model.



\begin{figure}[thp!]
\centering
\begin{subfigure}{.8\textwidth}
    \centering
    \includegraphics[width=\linewidth]{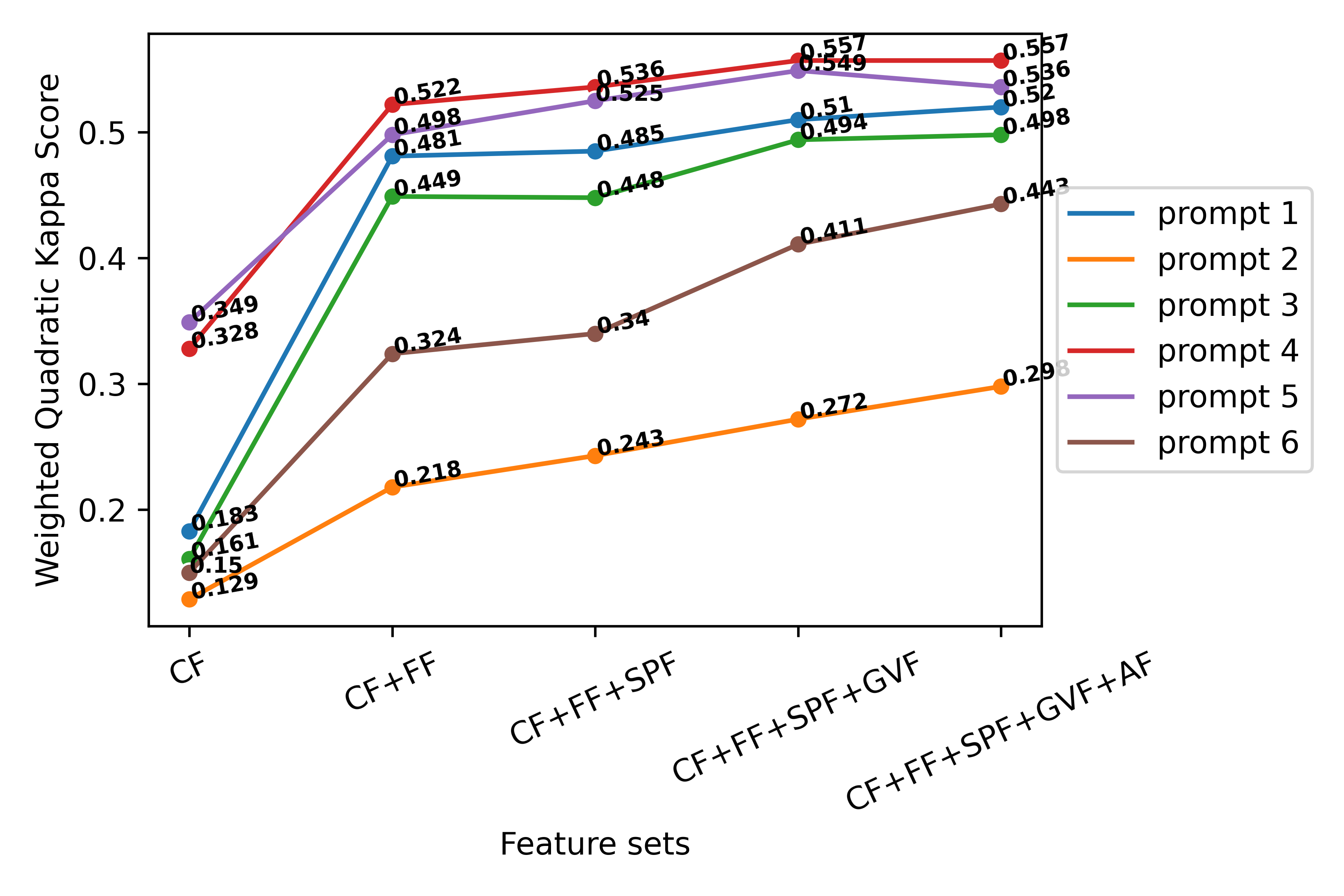}
\end{subfigure}
\caption{QWK for every prompt when a set of features is added one by one. Here, CF: Content Features, FF: Fluency Features, SPF: Superasegmantal Pronunciation Features, GVF: Grammar \& Vocabulary Features, AF: Audio Features. Here, the sequence in which feature groups are combined don't follow any relative ordering.}
 \label{fig:Ablation study 1}
\end{figure}

To get more insights into the impact of feature categories on the model's scoring ability, we set up two types of ablation studies. In the first one, we add feature categories (or sets) one by one to the base features set (which are content features in this case) and observe the change in QWK as we do so. Figure~\ref{fig:Ablation study 1} and Table~\ref{tab:Ablation study 1} depict the sharp increase in QWK for every prompt when the fluency features (FF) are introduced to the content features (CF). A similar significant increase is noticed for all the prompts when grammar and vocabulary features (GVF) are added to the previous feature set of content features, fluency features, and suprasegmental fluency features. When audio features (AF) are introduced, at last, we observe that QWK for prompts 1, 2, 4, and 6 increases while for the other prompts, the addition of audio features does not make any significant difference.

\begin{figure}[thp!]
\centering
\begin{subfigure}{.8\textwidth}
    \centering
    \includegraphics[width=\linewidth]{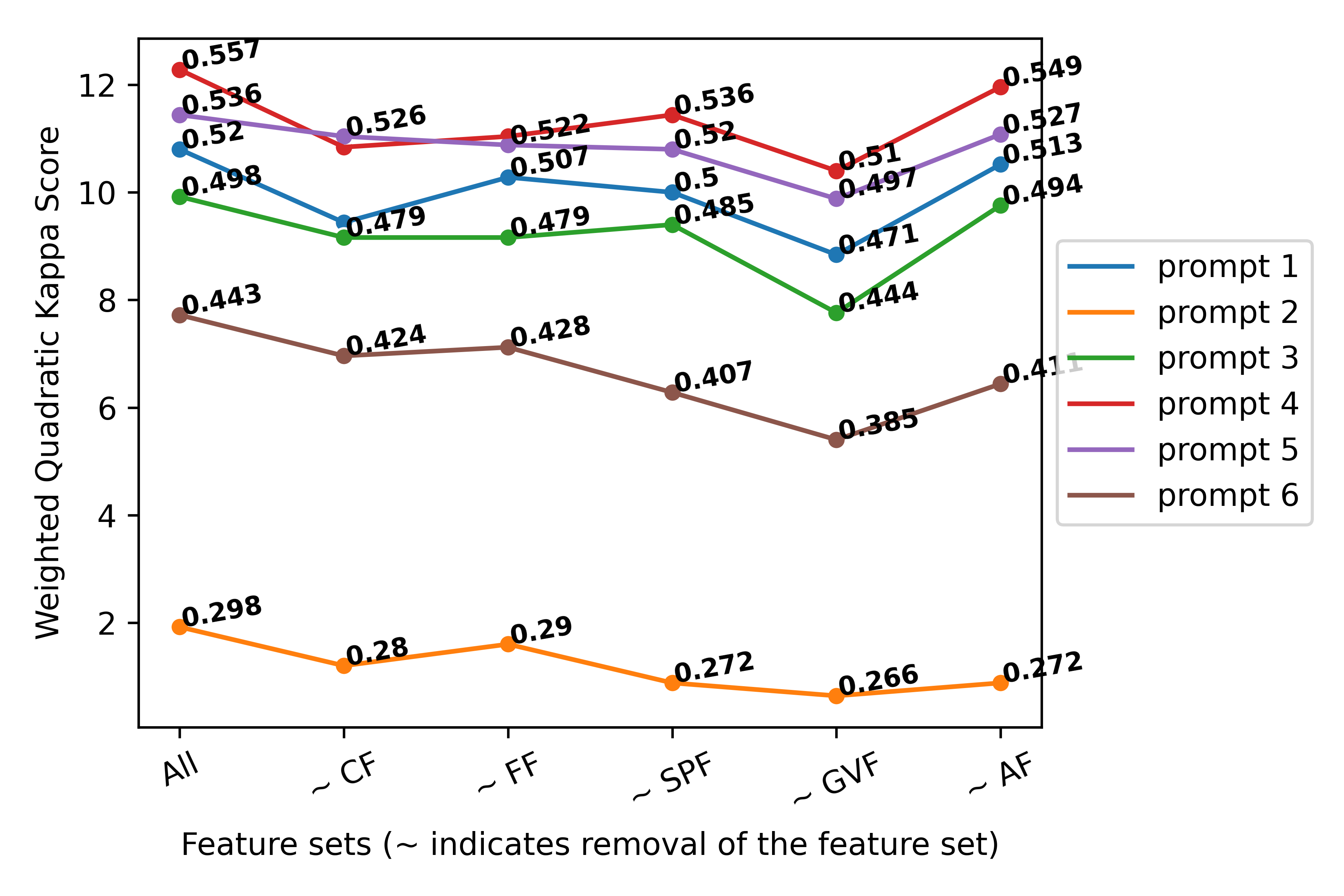}
\end{subfigure}%
\caption{QWK for every prompt when a set of features is dropped one by one while keeping other feature sets intact. Here, CF: Content Features, FF: Fluency Features, SPF: Superasegmantal Pronunciation Features, GVF: Grammar \& Vocabulary Features, AF: Audio Features.}
\label{fig:Ablation study 2}
\end{figure}

In the second ablation study, we remove one feature set at a time while keeping other feature sets intact. We report percentage change in the QWK metric in Table~\ref{tab:Ablation study 2} and also present the changed QWK in form of the graph in Figure~\ref{fig:Ablation study 2}. We observe that the impact of removing a feature set impacts each prompt differently. When content features (TF-IDF) were removed, prompts 1, 2, and 4 displayed drops in their QWK values by 6\% , while it was only approximately 4\% drop for prompts 3 and 6. Prompt 5 seems to have been affected the least by removing the content features. We believe this is because the prompt is relatively more open-ended and highly dependent on the individual's opinion. Grammar and vocabulary features (GVF) shows a significant impact on the QWK score when it is dropped as a feature set. The drop percentage ranges between 7-13\% for each prompt, thus confirming the importance of capturing grammatical construct and vocabulary usage. Fluency features (FF) and suprasegmental pronunciation features (SPF) tend to show have almost similar drop percentage range. The impact of dropping suprasegmental pronunciation features shows a maximum drop in QWK for prompts 2 and 6. Audio features (AF) have none to least impact on QWK for every prompt except for prompts 2 and 6. Our system does not do well for prompt 2 and prompt 6 when compared with the other prompts. One reason can be that the grammar, content, and pronunciation features were not able to capture the knowledge to understand why a test-taker was given a certain grade. Hence, the audio features and speech delivery metrics like rhythm and stress patterns are observed to be given more importance in scoring prompts 2 and 6. 

In both the ablation tests, the XGBoost regressor is trained from scratch. The first ablation study shows the contribution of a feature when added to the previous combination of the feature sets and shows the importance of the introduced feature set in improving QWK as it interacts with the features already present in the training set. The second ablation study questions the importance of a feature set by observing the magnitude of the drop in QWK as the feature set is removed during the training. With the feature set removed the feature interaction possibilities decreases within the training set. 

\begin{landscape}
    \vspace*{\fill}
    \begin{table}[h]
        \centering
        \setlength{\tabcolsep}{2pt}
          \caption{Quadratic Kappa (QWK) Score, Pearson correlation (\textit{r}) and Mean Squared Error (MSE) across prompts when each feature set is incrementally introduced to the previous feature set.}
         \label{tab:Ablation study 1}
          \begin{tabular}{*{17}{|c}}
          \hline
          \multirow{2}{*}{\textbf{XGBoost Regressor}} &
          \multicolumn{3}{c|}{\textbf{CF}} & 
          \multicolumn{3}{c|}{\textbf{CF + FF}}    & 
          \multicolumn{3}{c|}{\textbf{CF + FF + SPF}} & 
          \multicolumn{3}{c|}{\textbf{CF+FF+SPF+GVF}} & \multicolumn{3}{c|}{\textbf{\begin{tabular}[c]{@{}c@{}}CF+FF+SPF+GVF+AF\\ (Complete Feature Set)\end{tabular}}} \\
          \cline{2-16} 
          & \textbf{QWK}  & \textbf{r} & \textbf{MSE} 
          & \textbf{QWK}  & \textbf{r} & \textbf{MSE} 
          & \textbf{QWK}  & \textbf{r} & \textbf{MSE}  
          & \textbf{QWK}  & \textbf{r} & \textbf{MSE} 
          & \textbf{QWK}  & \textbf{r} & \textbf{MSE}                      \\ 
          \hline
          \textbf{Prompt 1}                                                                     & 0.183        & 0.236      & 0.317        
           & 0.481        & 0.515      & 0.224        
           & 0.485        & 0.521      & 0.223         
           & 0.510        & 0.531      & 0.220        
           & 0.520        & 0.550      & 0.211                             \\ 
           \hline
           \textbf{Prompt 2}                                                                 
           & 0.129        & 0.131       & 0.535        
           & 0.218        & 0.225       & 0.483        
           & 0.243        & 0.25        & 0.468         
           & 0.272        & 0.281       & 0.405        
           & 0.298        & 0.306       & 0.434                             \\ 
          \hline
          \textbf{Prompt 3}                                                                     & 0.161        & 0.202      & 0.586        
           & 0.449        & 0.466      & 0.463        
           & 0.448        & 0.467      & 0.459         
           & 0.494        & 0.508      & 0.436        
           & 0.498        & 0.512      & 0.430                             \\ 
          \hline
          \textbf{Prompt 4}                                                                     & 0.328        & 0.348      & 0.503        
           & 0.522        & 0.530      & 0.409        
           & 0.536        & 0.543      & 0.397         
           & 0.557        & 0.566      & 0.372       
           & 0.557        & 0.568      & 0.368                             \\ 
          \hline
          \textbf{Prompt 5}                                                                     & 0.349        & 0.365      & 0.487        
          & 0.498         & 0.510      & 0.401        
          & 0.525         & 0.538      & 0.382         
          & 0.549         & 0.560      & 0.366        
          & 0.536         & 0.549      & 0.371                             \\ 
          \hline
          \textbf{Prompt 6}                                                                     & 0.150        & 0.203      & 0.179        
           & 0.324        & 0.381      & 0.152        
           & 0.34         & 0.391      & 0.151         
           & 0.411        & 0.442      & 0.144        
           & 0.443        & 0.478      & 0.136                             \\ 
           \hline
           \end{tabular}
           
           \bigskip
            \caption{Percentage drop in QWK when the mentioned feature set is removed from the dataset, where CF: Content Features, FF: Fluency Features, SPF: Suprasegmental Pronunciation, GVP: Vocabulary and Grammar, AF: Acoustic Features and the symbol $\sim$ denotes removal of a feature set.}
             \label{tab:Ablation study 2}
            \begin{tabular}{|c|c|c|c|c|c|}
            \hline
            \multicolumn{1}{|l|}{} 
            & \multicolumn{1}{l|}{\textbf{$\sim$CF}} 
            & \multicolumn{1}{l|}{\textbf{$\sim$FF}} 
            & \multicolumn{1}{l|}{\textbf{$\sim$SPF}} 
            & \multicolumn{1}{l|}{\textbf{$\sim$GVF}} 
            & \multicolumn{1}{l|}{\textbf{$\sim$AF}} \\ \hline
            \textbf{Prompt 1}      
                    & -6.54                                  
                    & -2.50                                  
                    & -3.85                                   
                    & -9.43                                   
                    & -1.35                                  \\ \hline
            \textbf{Prompt 2}      
                    & -6.05                                  
                    & -2.69                                  
                    & -8.73                                   
                    & -10.74                                  
                    & -8.73                                  \\ \hline
            \textbf{Prompt 3}      
                    & -3.82                                  
                    & -3.82                                  
                    & -2.62                                   
                    & -10.85                                  
                    & -0.81                                  \\ \hline
            \textbf{Prompt 4}      
                    & -6.47                                  
                    & -5.57                                  
                    & -3.78                                   
                    & -7.72                                   
                    & -1.44                                  \\ \hline
            \textbf{Prompt 5}      
                    & -1.87                                  
                    & -2.62                                  
                    & -2.99                                   
                    & -7.28                                   
                    & -1.68                                  \\ \hline
            \textbf{Prompt 6}      
                    & -4.29                                  
                    & -3.39                                  
                    & -8.13                                   
                    & -13.10                                  
                    & -7.23                                  \\ \hline
            \end{tabular}
        \end{table}
        \vspace*{\fill}
        \end{landscape}

\subsection{Feature Impact on the Model's Outcome}
In the feature section of the paper, we calculated the Pearson correlation between each feature and the human-rated score. Features like speaking rate, articulation rate, total text complexity features, wordsyll2, wordtypes, etc. show a positive correlation with the scores given by expert-human raters, while features silence features like filled\_pause\_rate, general\_silence, SilenceRate1, long\_silence\_deviation, etc. are negatively correlated with the scores. Despite knowing the correlation of individual features with the scores, we do not exactly know at what feature value the probability of getting a higher score increases or decreases. Thus, we select a few important features and utilize model-agnostic methods of model interpretation like Partial Dependence Plots and Shapley values to dig a little deep into knowing how the linguistic features we calculated to impact the scoring process. We use Prompt 4 for further analysis in this section because of its high scores as presented by the evaluation metrics, and has more number of classes. The same results can be translated to the other prompts as well.

\subsubsection{Partial Dependence Plots for Linguistic Features}
Partial Dependence Plot (PDP plot) \citep{pdpth} is used to show the type of relationship (linear, monotonic, or more complex) between a feature and the target, and the effect of one or two features on the predicted outcome of a machine learning model. For regression, the partial dependence function is defined as:

\begin{equation}
\label{Eq:pd}
  \hat{f}_{X_{s}}(X_{s}) = \int\hat{f}(X_{s}, X_{c})dP(X_{c})
\end{equation}

Here, $X_{s}$ denotes the feature for which the partial dependence function is to be calculated, and $X_{c}$ denotes the other features in the training set. $\hat{f}$ denotes the machine learning model that utilizes both $X_{s}$ and $X_{c}$ which makes up for the total feature space $X$. The partial dependence function marginalize the output of the machine learning model over the distribution present in the feature set $X_{c}$. This helps in showing the relationship between the predicted outcome and the feature of which we want to study the effect of the model's predictions.

Similar to \cite{pdpaplication}, we utilize these plots to understand the relationship of individual features with the target as learned by the trained model. For this, we handpick a few features from the features that were consistently marked important by the model as shown in the feature importance graphs. 
\begin{figure}
\centering
    \includegraphics[width=12cm]{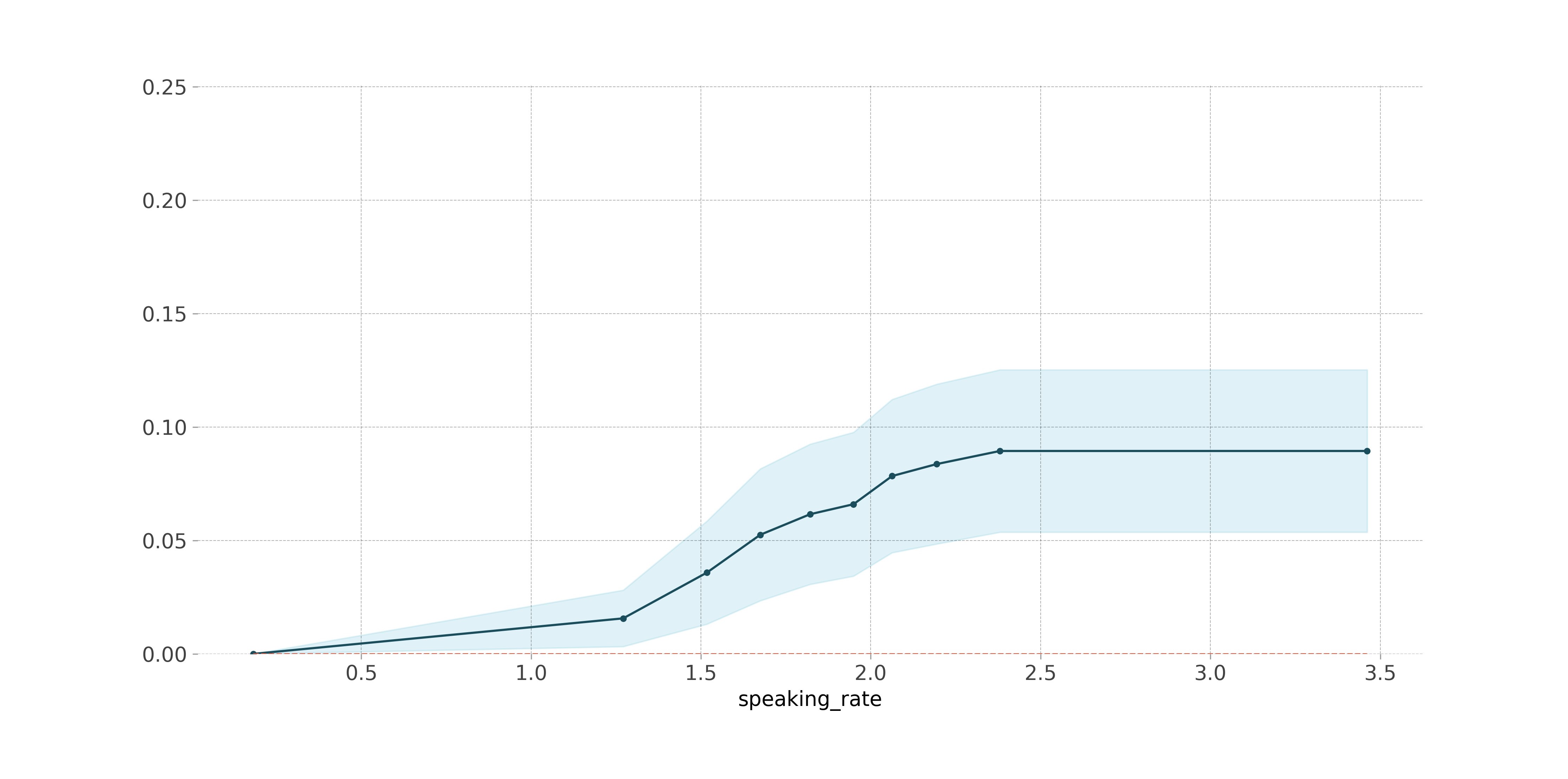}
    \caption{PDP plot for speaking rate.}
    \includegraphics[width=12cm]{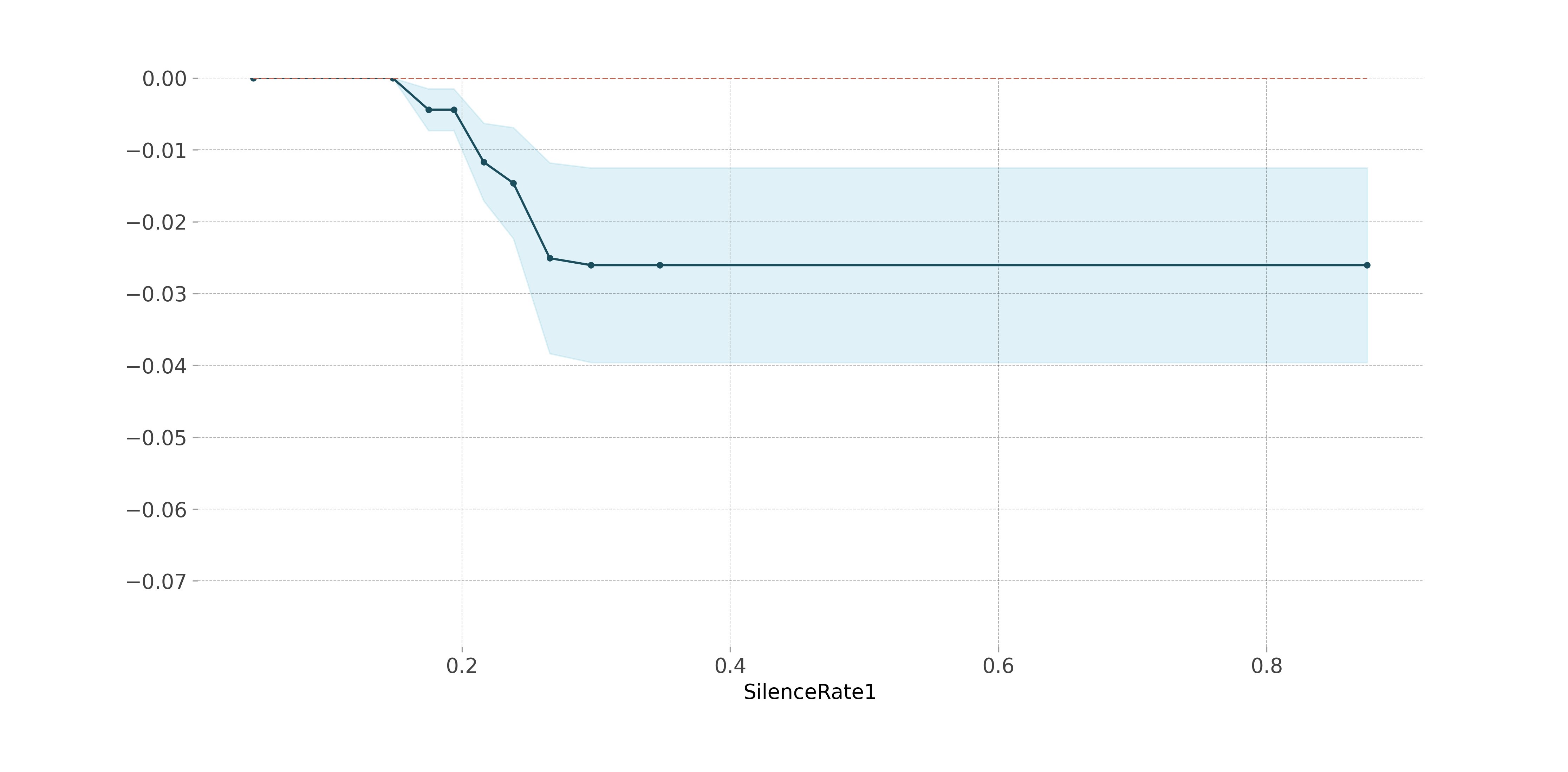}
    \caption{PDP plot for silence rate 1.}
    \includegraphics[width=12cm]{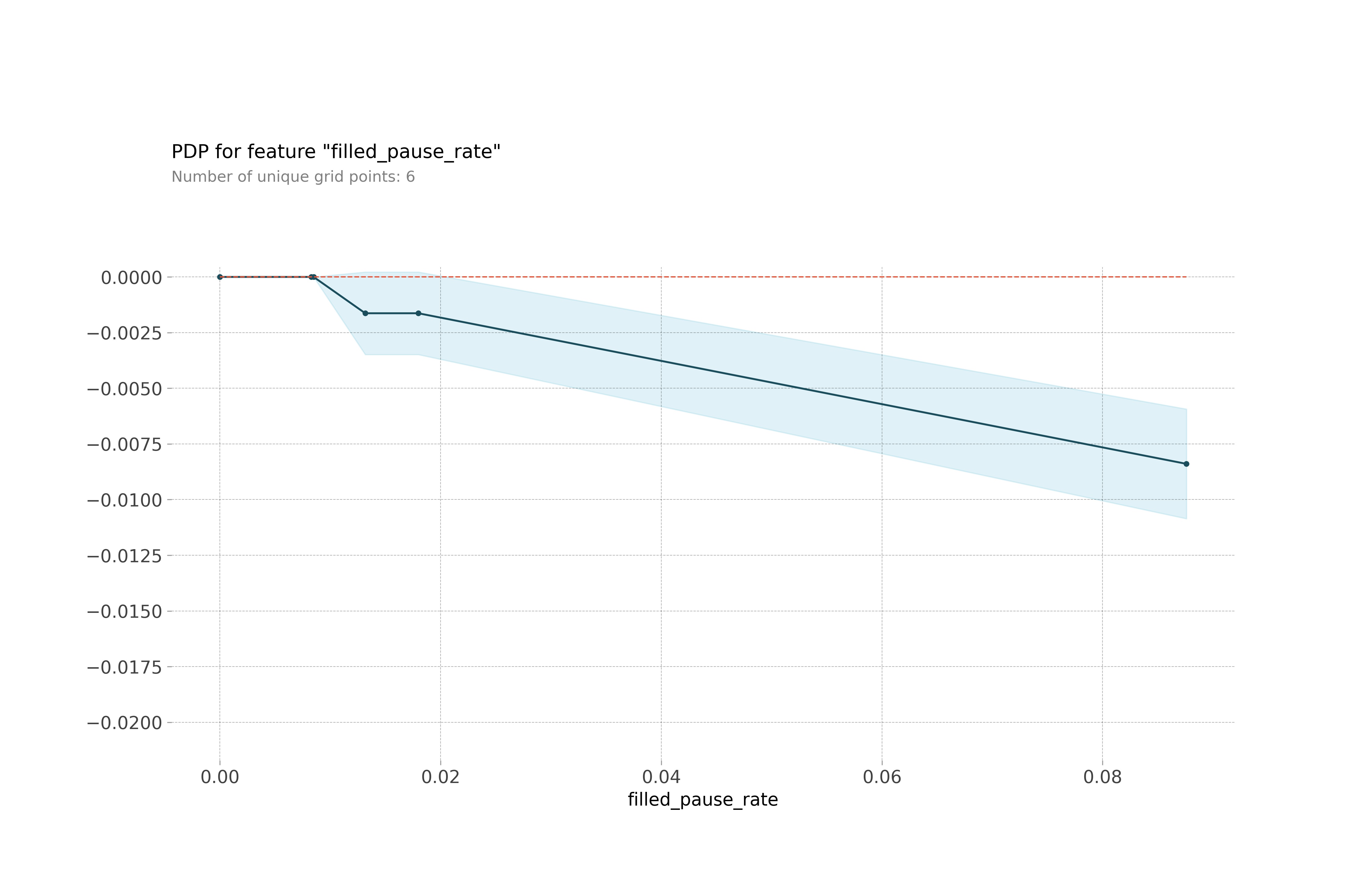}
    \caption{PDP plot for filled pause rate.}
    \label{fig:PDP Plots for Fluency Features}
\end{figure}

\textbf{Fluency Features}: We evaluate some of the features from the fluency feature category like speaking rate, rate of filled pauses, and silence rate 1. The speaking rate is the number of words spoken per second. Intuitively, a proficient speaker will speak more words per second but if this value is beyond a certain point the speech may become unintelligible. The partial dependence plot (PD Plot) for speech rate on our dataset shows a steady increase in the probability of getting a higher score from up till the speaking rate value of 1.35 (approx), followed by a sharp increase up till the speaking rate value of 2.4 (approx). Beyond a speaking rate of 2.4, the probability of getting a higher score does not really get influenced. This shows that for the dataset of L2 Filipino Speakers, the 2 to almost 3 words spoken per second is considered a quality of a proficient speaker. Silence rate 1 is the measure of the number of silences divided by total words spoken in the response time. The high value of silence rate 1 indicates the presence of more silent segments in the response time. As the PD plot for silence rate 1 depicts, the higher the value of this feature, the lower the probability of getting a higher grade. However, this trend is only supported between the approx ranges of 0.18 to 0.27 number of silences per word. Beyond 0.27 number of silences per word, this feature value does not help much in determining a lower score. Like silence rate 1, the rate of the filler pauses has a negative correlation with the scores given by the human raters. Filler pauses or filler words are the usage of sounds like ``uh", ``umm", ``err", etc. which introduces disfluency to the speech. The more the amount of such word in the spoken content, the lesser the probability of getting a higher grade. Similar intuition is supported by the PD plot for the filled pause rate which shows the continuous decline as the value of pause words increases in the spoken content. Studying these fluency features confirms the intuition learned by the model is as per the intuition followed by the expert human raters.

\begin{figure}
\centering
    \includegraphics[width=12cm]{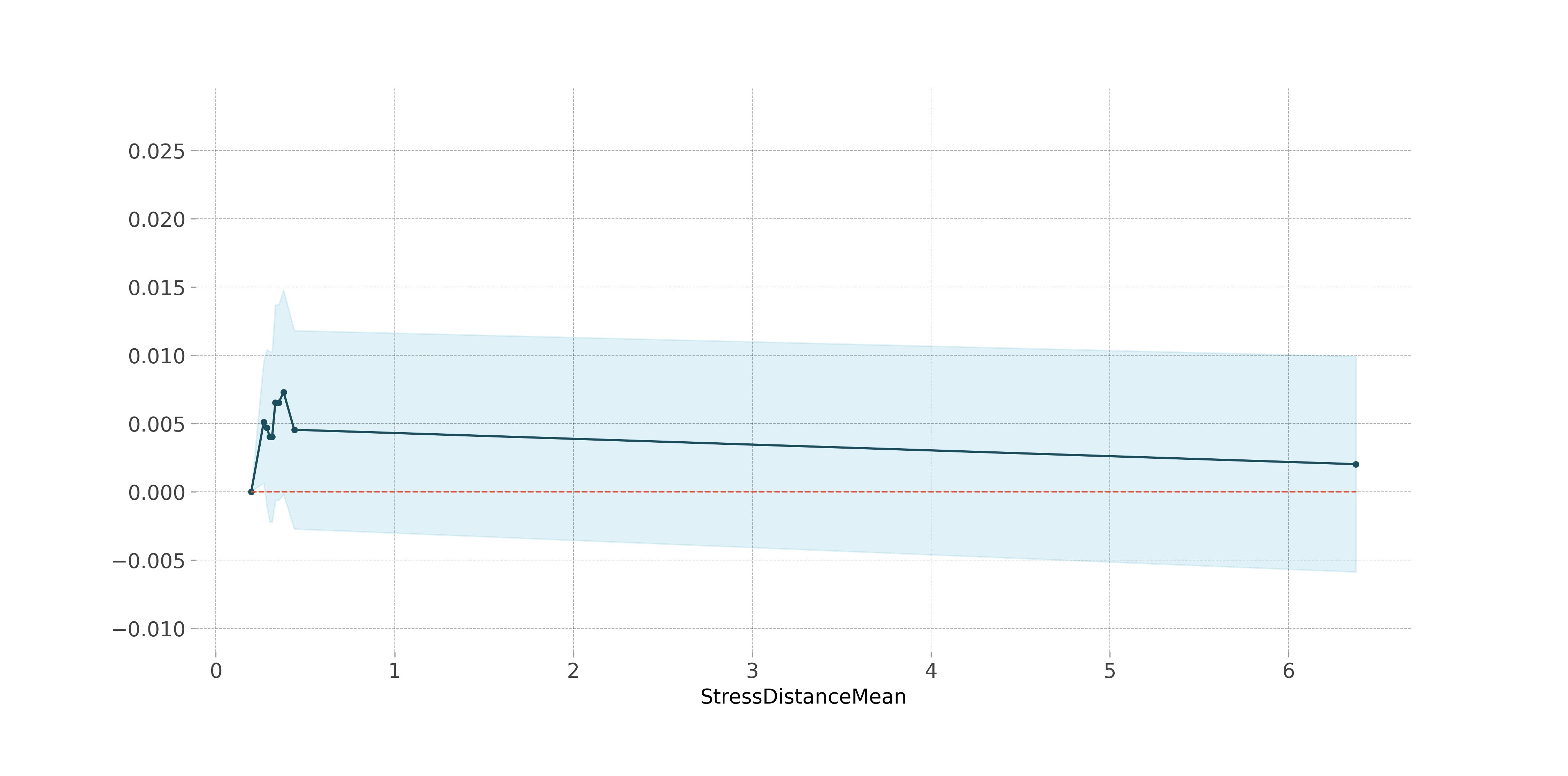}
    \caption{PDP plot for StressDistanceMean.}
    \includegraphics[width=12cm]{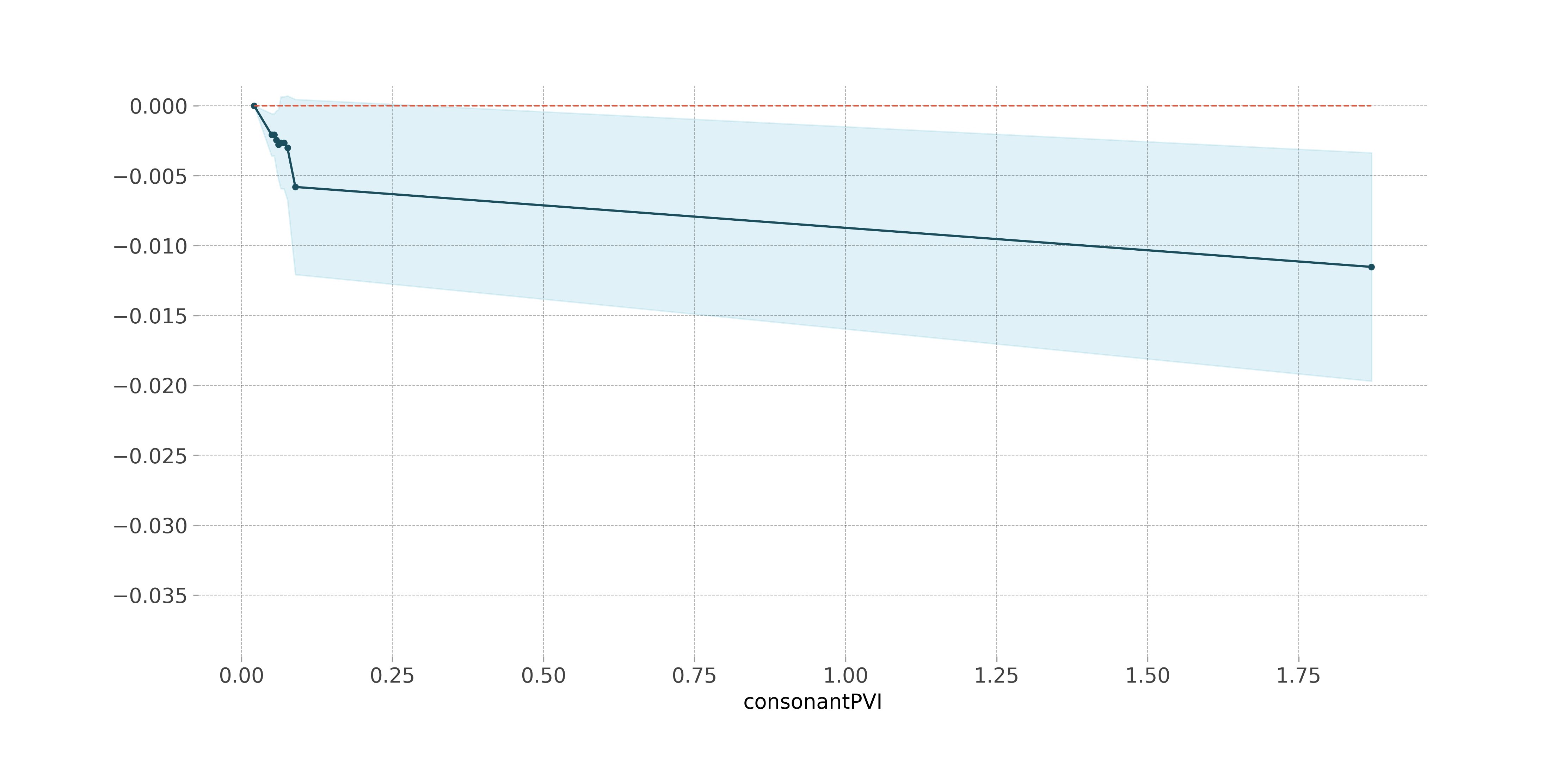}
    \caption{PDP plot for consonantPVI.}
    \label{fig:PDP Plots for Pronunciation Features}
\end{figure}

\textbf{Suprasegmental Pronunciation Features}: Rhythmic and stress-based features capture the stress patterns and occurrences of vowels, consonants, and syllables in the spoken content. The features we will discuss are StressDistanceMean and consonantPVI. StressDistanceMean is the mean distance between stressed syllables in seconds. As given in Table~\ref{Tab:stressbased}, this feature has a negative correlation with the human-rated score. Every syllable has at least one stressed phoneme. The decrease in the distance between stressed syllables can be indicative of either using high vocabulary words as they tend to have more stressed phonemes or simply using fewer pauses when speaking. As suggested by PD Plots, the increased value of this feature is responsible for getting slightly lower scores. ConsonantPVI is the Raw Pairwise Variability Index (PVI) of consonantal phonemes. This feature captures the patterns of successive consonantal intervals. As per the correlation from Table~\ref{Tab:rhythmbased}, this feature has a negative correlation with the score. The PD plot of this feature confirms the negative correlation and shows a steady decline as the feature values increases from 0.1 to 1.75 and beyond.

\begin{figure}
\centering
    \includegraphics[width=12cm]{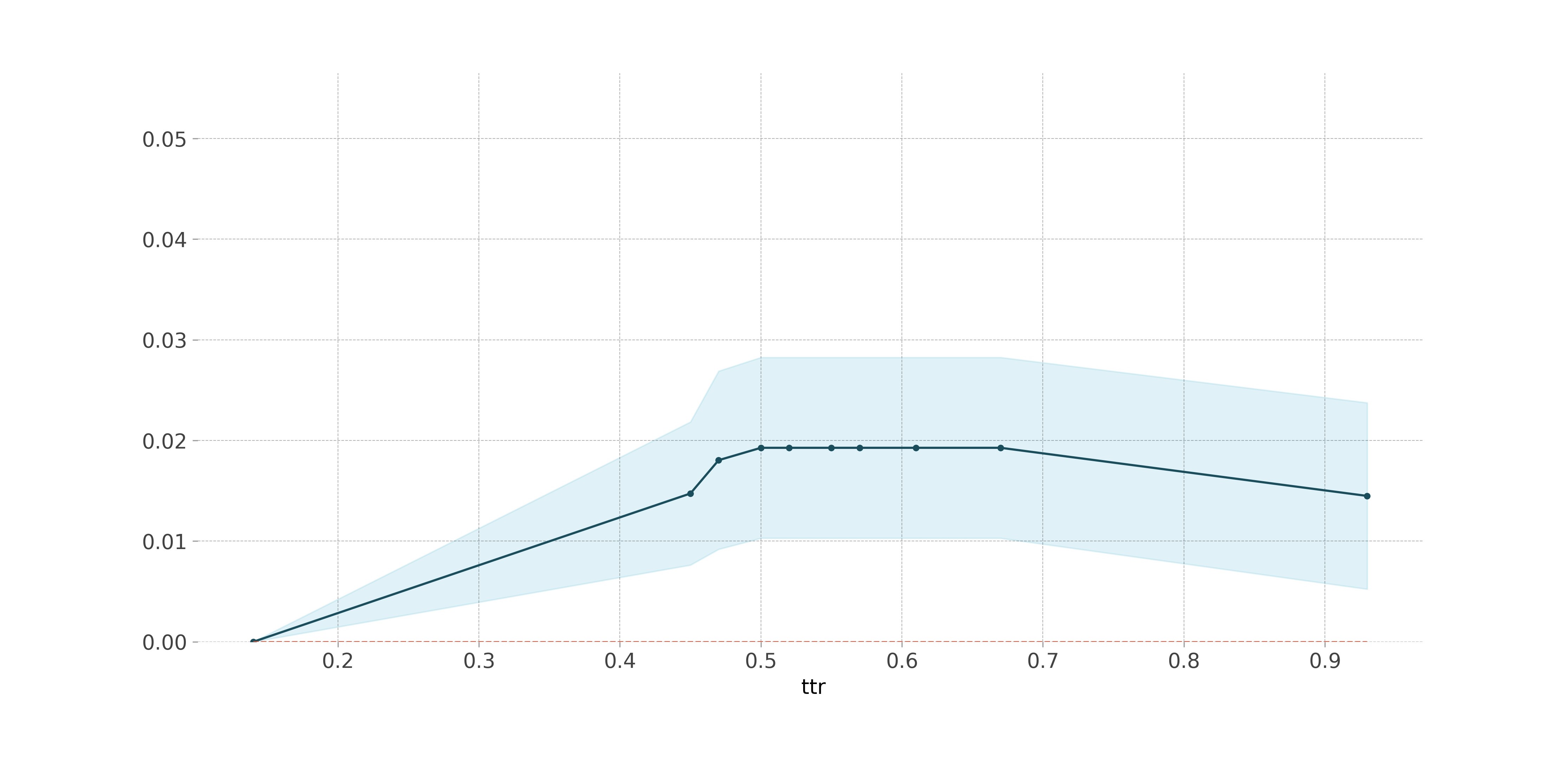}
    \caption{PDP plot for TTR (Type to Token Ratio).}
    \includegraphics[width=12cm]{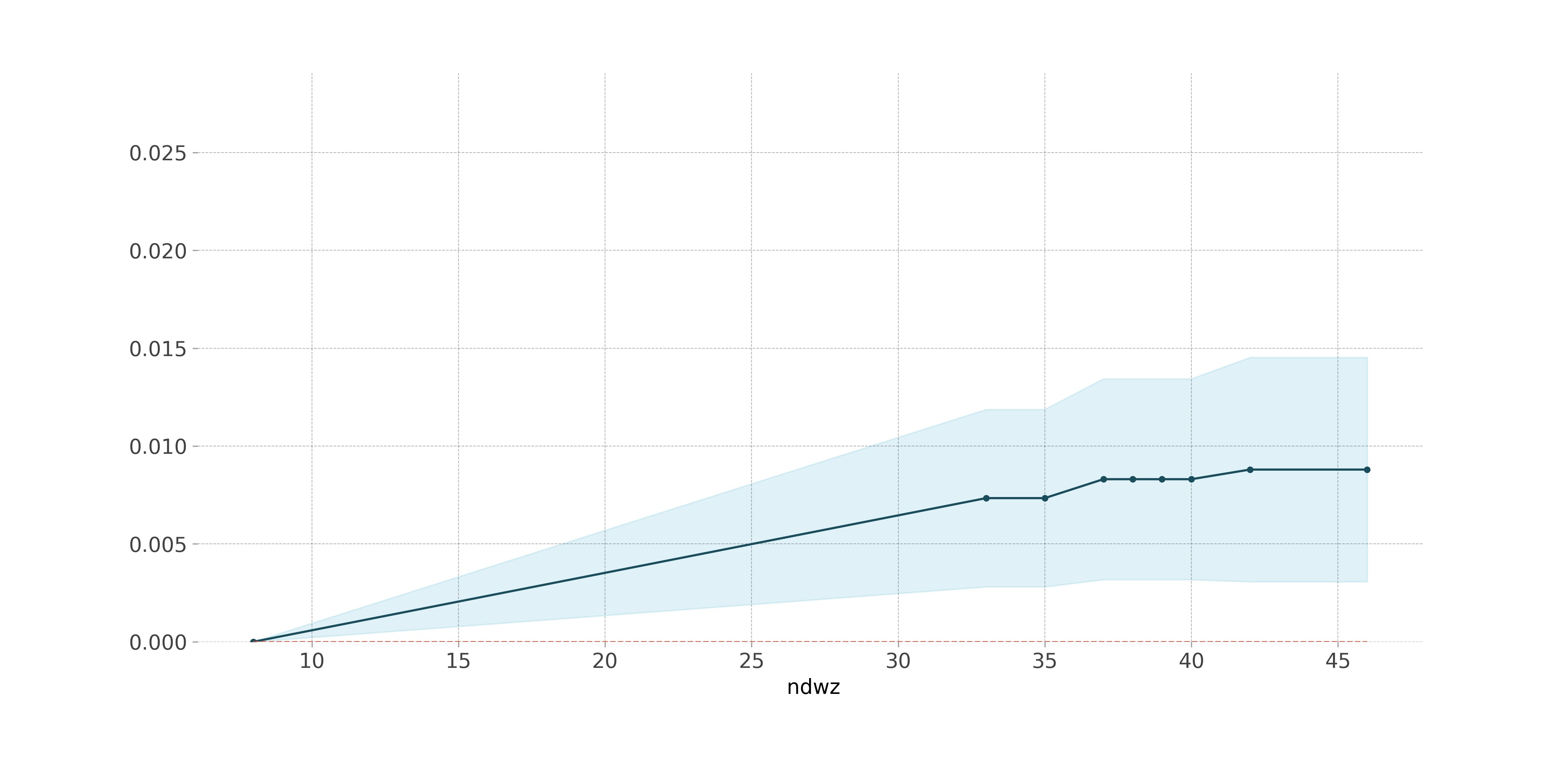}
    \caption{PDP plot for NDWZ (Number of different words from first 50 words).}
    \includegraphics[width=12cm]{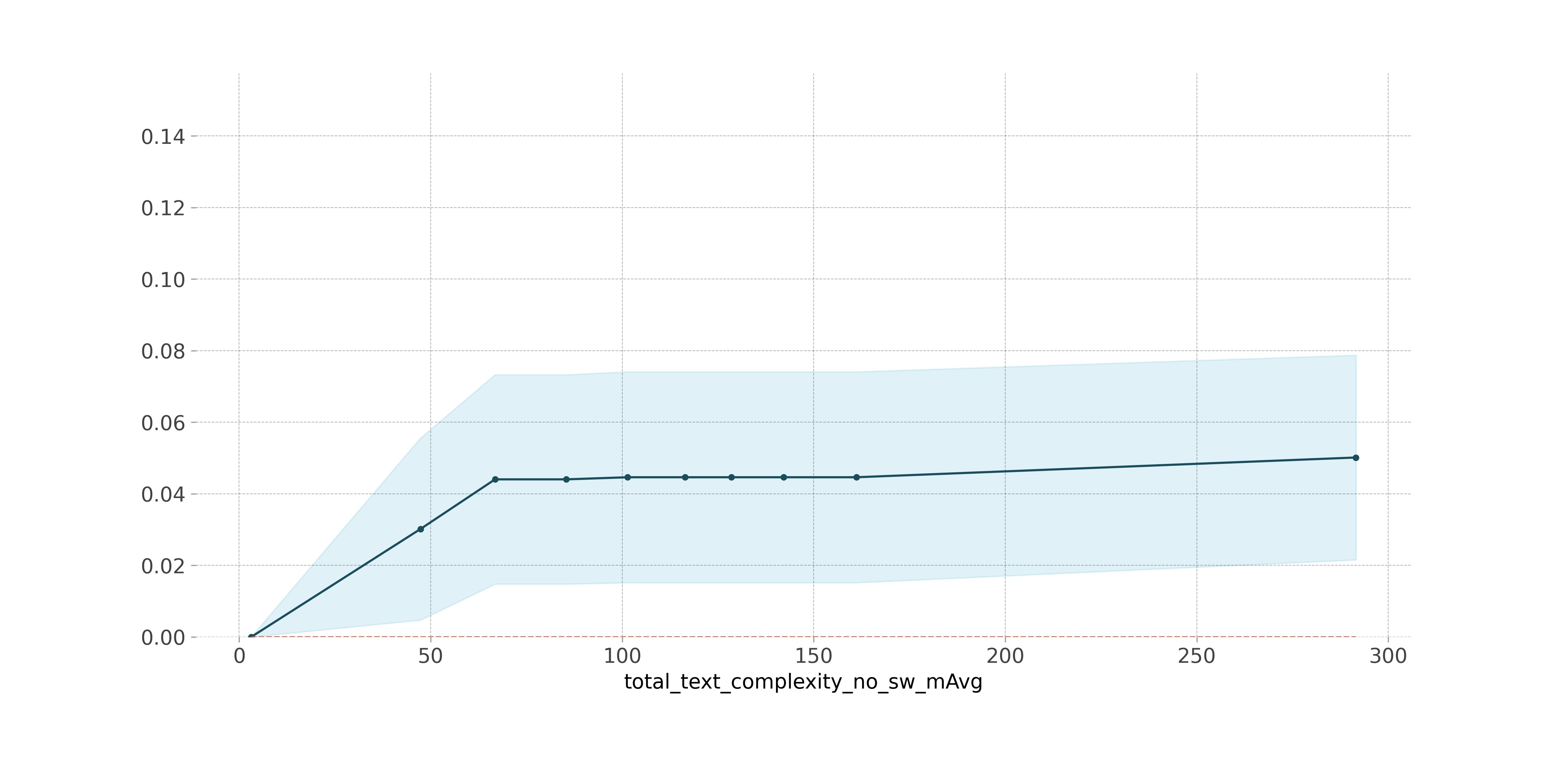}
    \caption{PDP plot for total\_text\_complexity\_no\_sw\_mAvg.}
    \label{fig:PDP Plots for Grammar and Vocabulary Features}
\end{figure}

\textbf{Grammar and Vocabulary Features}: We analyze the type-to-token ratio, total text complexity when stop words are removed and the number of different words from the first 50 words. TTR or type-to-token ratio is obtained by dividing the number of different words divided by the number of total utterances spoken in the response. If the value of this feature is high ,it indicated a high degree of lexical variation and when it is low it indicates vice-versa. TTR shows a positive correlation with our dataset as mentioned in Table ~\ref{Tab:lexicalcomplexity}. With the increase in lexical variation in the spoken response, the test-taker tends to get a higher score. The PD Plot for TTR shows a steady increase in the probability of getting a higher score until the TTR value of 0.45 approx and shows a sharp increase up to 0.5. The probability of getting a higher score because of the higher value of TTR gets negatively impacted beyond the TTR value of 0.68, thus showing a sharp decline in the slope. NDWZ is the number of different words from the first 50 words. The PD Plot suggests the increase in the probability of getting higher scores as the value of this feature increases. Total\_text\_complexity\_no\_sw\_mAvg is the total complexity of words in the response when all the stop words are removed from it. The higher vocabulary words tend to have higher complexity. Thus, here, the higher value of text complexity indicates a higher probability of getting a high score. The PD Plot for this feature indicates a sharp rise in the probability of getting a high score from 0 (audio responses where no word was spoken or out of vocabulary word was spoken) to 65 (approx). Then beyond the complexity score of 155, it again shows a steady increase in the probability of getting a higher score.

\subsubsection{Explaining the model via SHAP Values}
Shapley value \citep{shapth} is a concept borrowed from coalition game theory. Its ability to capture the effect of the grouping feature makes it a useful application to the feature selection problem. Shapley value is the average marginal contribution of a feature value in enhancing the accuracy of models' prediction across all the possible coalitions. 

\begin{figure}[h]
\centering
    \includegraphics[width=10cm]{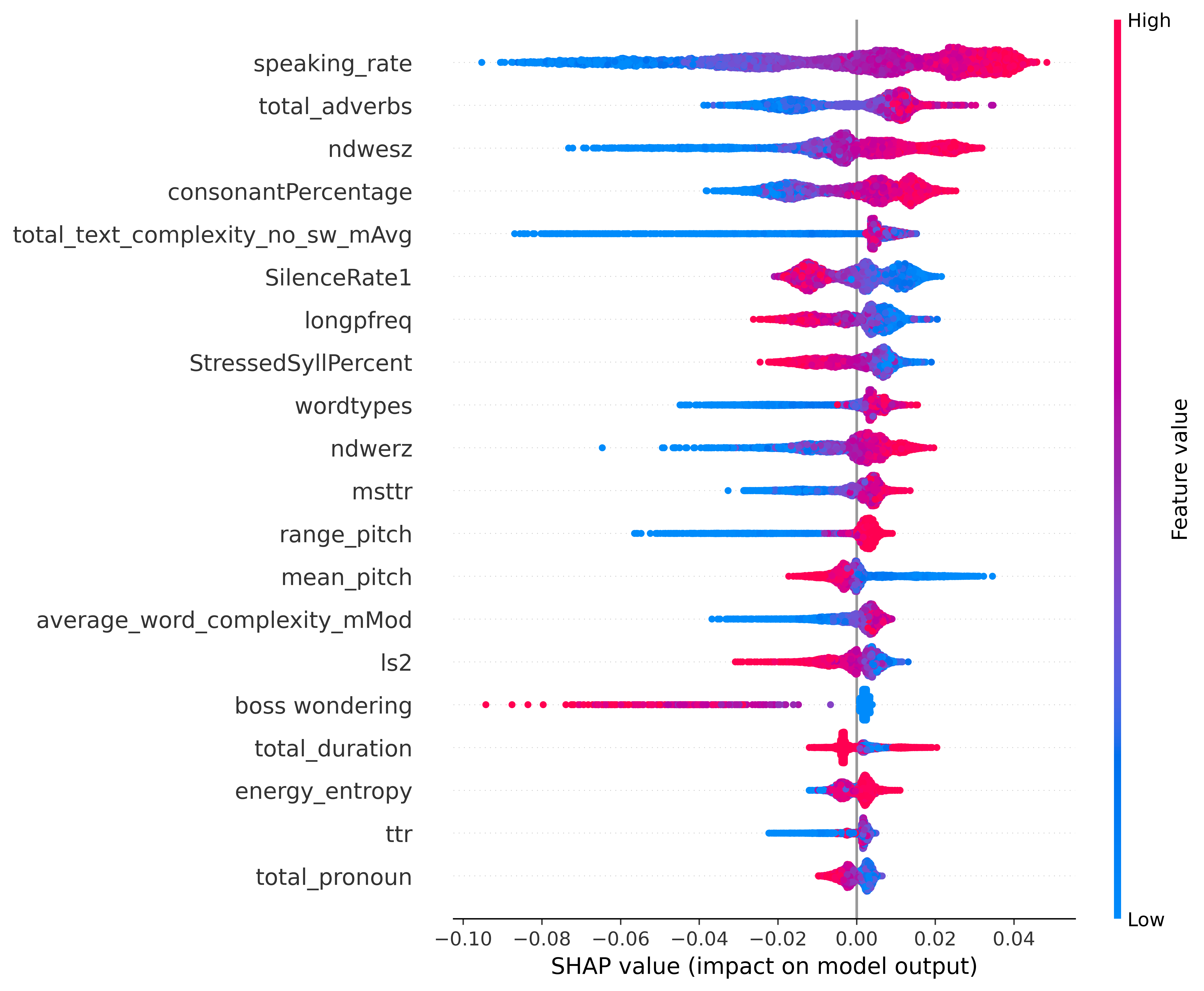}
    \caption{SHAP Summary Plot: Impact of each feature on the predictions generated by the model.}
    \label{fig:Shapsummaryplot}
\end{figure}

In this paper, we utilize Shapley Additive exPlanations \citep{shap} instead. It is based on Shapley values and helps in the explanation of individual predictions. It computes the contribution of each feature for making a certain prediction given an data point from the dataset and thus explains why a certain prediction was made. Additionally, it also gives an average impact of features on the models' prediction ability. In this paper, we use of SHAP Summary Plot which combines feature importance with feature effects. Every data point on this graph denotes a Shapley value for both, a feature and an instance. The x-axis of this plot denotes Shapley values and the y-axis shows the features ordered according to their importance as perceived by the trained model. The overlapping data points give us the sense of the distribution of that feature value and the color shows if the value of a feature is high (red) or low (blue) for a specific data point.  

In the summary plot for Prompt 4 given in Figure~\ref{fig:Shapsummaryplot}, the density of data distribution in red and maximum on the positive side of the x-axis for speaking\_rate, total\_adverb count, ndwesz, consonant percentage, etc. which shows a positive correlation of these features with the score. Similarly, SilenceRate1, longpfreq, StressedSyllPercent etc., show a negative correlation. The interpretation that can be drawn from the graph is that for speakers with a higher speaking rate, high variation is the usage of different words, high vocabulary words, and small intervals of pauses will have a high probability of scoring a higher grade.


\section{Conclusion and Future Work}
\label{sec:conclusion}
In this paper, we introduced the SLTI SOPI Automated Scoring System and performed thorough experiments to investigate how the linguistic features impact the speech scoring model. First, we show that formulating a speech scoring task as a regression analysis performs better than classification. Through thorough experiments, the XGBoost regressor was selected as the best performing model. Through ablation studies, we show the importance of each feature set across different prompts revealing the relationship between the feature set and the model trained for that prompt. Additionally, we used model-agnostic model interpretability methods--- Partial Dependence Plots and Shapley values, to reveal the impact of the selected linguistic features on the model's scoring ability. We showed the specific value of the linguistic feature and explained the trend of scoring with the change in that feature value. In doing so, we confirmed the ability of our model to pick up essential cues from these linguistic features to draw intuitions similar to the expert-human raters.

The next version of the SLTI SOPI Automated Scoring System, will include many more important sets of features required to build a complete solution to the speech scoring task like segmental pronunciation features, grammatical accuracy score, and coherence features. We also plan to make use of deep learning to further explore the problem domain. All these experiments we plan to do in the near future will be done with the intention to make the speech grading model more interpretable.  


%


\section{Declarations}

\section*{Funding}
Yaman Kumar Singla is the recipient of Google and Prime Minister PhD fellowships and would like to thank Google and CII-SERB for supporting him. Rajiv Ratn Shah is partly supported by the Infosys Center for AI and Center for Design and New Media at IIIT-Delhi and would like to thank them.

%
\section*{Conflict of interest}

Not applicable.

\section*{Availability of data and material}

Not applicable.

\section*{Code availability}

Not applicable.

\bibliographystyle{spbasic}      
\bibliography{bibliography}   

\begin{thebibliography}{60}
\providecommand{\natexlab}[1]{#1}
\providecommand{\url}[1]{{#1}}
\providecommand{\urlprefix}{URL }
\expandafter\ifx\csname urlstyle\endcsname\relax
  \providecommand{\doi}[1]{DOI~\discretionary{}{}{}#1}\else
  \providecommand{\doi}{DOI~\discretionary{}{}{}\begingroup
  \urlstyle{rm}\Url}\fi
\providecommand{\eprint}[2][]{\url{#2}}

\bibitem[{Ai and Lu(2010)}]{tool}
Ai H, Lu X (2010) A web-based system for automatic measurement of lexical
  complexity. In: 27th Annual Symposium of the Computer-Assisted Language
  Consortium (CALICO-10). Amherst, MA. June, pp 8--12

\bibitem[{Amodei et~al.(2016)Amodei, Ananthanarayanan, Anubhai, Bai,
  Battenberg, Case, Casper, Catanzaro, Chen, Chrzanowski, Coates, Diamos,
  Elsen, Engel, Fan, Fougner, Hannun, Jun, Han, LeGresley, Li, Lin, Narang, Ng,
  Ozair, Prenger, Qian, Raiman, Satheesh, Seetapun, Sengupta, Wang, Wang, Wang,
  Xiao, Xie, Yogatama, Zhan, and Zhu}]{DBLP:conf/icml/AmodeiABCCCCCCD16}
Amodei D, Ananthanarayanan S, Anubhai R, Bai J, Battenberg E, Case C, Casper J,
  Catanzaro B, Chen J, Chrzanowski M, Coates A, Diamos G, Elsen E, Engel J, Fan
  L, Fougner C, Hannun AY, Jun B, Han T, LeGresley P, Li X, Lin L, Narang S, Ng
  AY, Ozair S, Prenger R, Qian S, Raiman J, Satheesh S, Seetapun D, Sengupta S,
  Wang C, Wang Y, Wang Z, Xiao B, Xie Y, Yogatama D, Zhan J, Zhu Z (2016) Deep
  speech 2 : End-to-end speech recognition in english and mandarin. In: ICML,
  pp 173--182

\bibitem[{Ardila et~al.(2020)Ardila, Branson, Davis, Kohler, Meyer, Henretty,
  Morais, Saunders, Tyers, and Weber}]{ardila-EtAl:2020:LREC}
Ardila R, Branson M, Davis K, Kohler M, Meyer J, Henretty M, Morais R, Saunders
  L, Tyers F, Weber G (2020) Common voice: A massively-Â­multilingual speech
  corpus. In: Proceedings of The 12th Language Resources and Evaluation
  Conference, European Language Resources Association, Marseille, France, pp
  4218--4222, \urlprefix\url{https://www.aclweb.org/anthology/2020.lrec-1.520}

\bibitem[{Bachman et~al.(1996)Bachman, Palmer et~al.}]{bachman1996language}
Bachman LF, Palmer AS, et~al. (1996) Language testing in practice: Designing
  and developing useful language tests, vol~1. Oxford University Press

\bibitem[{Boersma and Van~Heuven(2001)}]{boersma2001speak}
Boersma P, Van~Heuven V (2001) Speak and unspeak with praat. Glot International
  5(9/10):341--347

\bibitem[{Burstein and Chodorow(1999)}]{burstein-chodorow-1999-automated}
Burstein J, Chodorow M (1999) Automated essay scoring for nonnative {E}nglish
  speakers. In: Computer Mediated Language Assessment and Evaluation in Natural
  Language Processing,
  \urlprefix\url{https://www.aclweb.org/anthology/W99-0411}

\bibitem[{Chen and Zechner(2011{\natexlab{a}})}]{rhythm}
Chen L, Zechner K (2011{\natexlab{a}}) Applying rhythm features to
  automatically assess non-native speech. In: Twelfth Annual Conference of the
  International Speech Communication Association

\bibitem[{{Chen} et~al.(2018){Chen}, {Tao}, {Ghaffarzadegan}, and
  {Qian}}]{8462562}
{Chen} L, {Tao} J, {Ghaffarzadegan} S, {Qian} Y (2018) End-to-end neural
  network based automated speech scoring. In: 2018 IEEE International
  Conference on Acoustics, Speech and Signal Processing (ICASSP), pp 6234--6238

\bibitem[{Chen et~al.(2018)Chen, Zechner, Yoon, Evanini, Wang, Loukina, Tao,
  Davis, Lee, Ma, Mundkowsky, Lu, Leong, and Gyawali}]{speechrater}
Chen L, Zechner K, Yoon SY, Evanini K, Wang X, Loukina A, Tao J, Davis L, Lee
  CM, Ma M, Mundkowsky R, Lu C, Leong CW, Gyawali B (2018) Automated scoring of
  nonnative speech using the speechratersm v. 5.0 engine. ETS Research Report
  Series 2018, \doi{10.1002/ets2.12198}

\bibitem[{Chen and Zechner(2011{\natexlab{b}})}]{syntactic}
Chen M, Zechner K (2011{\natexlab{b}}) Computing and evaluating syntactic
  complexity features for automated scoring of spontaneous non-native speech.
  In: Proceedings of the 49th Annual Meeting of the Association for
  Computational Linguistics: Human Language Technologies, Association for
  Computational Linguistics, Portland, Oregon, USA, pp 722--731,
  \urlprefix\url{https://www.aclweb.org/anthology/P11-1073}

\bibitem[{{Council of Europe. Council for Cultural Co-operation. Education
  Committee. Modern Languages Division}(2001)}]{council2001common}
{Council of Europe Council for Cultural Co-operation Education Committee Modern
  Languages Division} (2001) Common European Framework of Reference for
  Languages: learning, teaching, assessment. Cambridge University Press

\bibitem[{Craighead et~al.(2020)Craighead, Caines, Buttery, and
  Yannakoudakis}]{craighead-etal-2020-investigating}
Craighead H, Caines A, Buttery P, Yannakoudakis H (2020) Investigating the
  effect of auxiliary objectives for the automated grading of learner {E}nglish
  speech transcriptions. In: Proceedings of the 58th Annual Meeting of the
  Association for Computational Linguistics, Association for Computational
  Linguistics, Online, pp 2258--2269, \doi{10.18653/v1/2020.acl-main.206},
  \urlprefix\url{https://www.aclweb.org/anthology/2020.acl-main.206}

\bibitem[{Dikli and Bleyle(2014)}]{DIKLI20141}
Dikli S, Bleyle S (2014) Automated essay scoring feedback for second language
  writers: How does it compare to instructor feedback? Assessing Writing 22:1
  -- 17, \doi{https://doi.org/10.1016/j.asw.2014.03.006},
  \urlprefix\url{http://www.sciencedirect.com/science/article/pii/S1075293514000221}

\bibitem[{Ding et~al.(2020)Ding, Riordan, Horbach, Cahill, and
  Zesch}]{ding2020don}
Ding Y, Riordan B, Horbach A, Cahill A, Zesch T (2020) Don’t take
  “nswvtnvakgxpm” for an answer--the surprising vulnerability of automatic
  content scoring systems to adversarial input. In: Proceedings of the 28th
  International Conference on Computational Linguistics, pp 882--892

\bibitem[{Friedman(1991)}]{pdpth}
Friedman JH (1991) Multivariate adaptive regression splines. The annals of
  statistics pp 1--67

\bibitem[{Grover et~al.(2020{\natexlab{a}})Grover, Bamdev, Kumar, Hama, and
  Shah}]{grover2020audino}
Grover MS, Bamdev P, Kumar Y, Hama M, Shah RR (2020{\natexlab{a}}) audino: {A}
  modern annotation tool for audio and speech. CoRR abs/2006.05236,
  \urlprefix\url{https://arxiv.org/abs/2006.05236}, \eprint{2006.05236}

\bibitem[{Grover et~al.(2020{\natexlab{b}})Grover, Kumar, Sarin, Vafaee, Hama,
  and Shah}]{grover2020multi}
Grover MS, Kumar Y, Sarin S, Vafaee P, Hama M, Shah RR (2020{\natexlab{b}})
  Multi-modal automated speech scoring using attention fusion. arXiv preprint
  arXiv:200508182 \urlprefix\url{https://arxiv.org/abs/2005.08182}

\bibitem[{Hsieh et~al.(2019{\natexlab{a}})Hsieh, Zechner, and
  Xi}]{hsieh2019features}
Hsieh CN, Zechner K, Xi X (2019{\natexlab{a}}) Features measuring fluency and
  pronunciation. In: Automated Speaking Assessment, Routledge, pp 101--122

\bibitem[{Hsieh et~al.(2019{\natexlab{b}})Hsieh, Zechner, and Xi}]{chflu}
Hsieh CN, Zechner K, Xi X (2019{\natexlab{b}}) Features measuring fluency and
  pronunciation. Automated Speaking Assessment: Using Language Technologies to
  Score Spontaneous Speech p 101

\bibitem[{Johan~Berggren et~al.(2019)Johan~Berggren, Rama, and
  {\O}vrelid}]{classreg}
Johan~Berggren S, Rama T, {\O}vrelid L (2019) Regression or classification?
  automated essay scoring for {N}orwegian. In: Proceedings of the Fourteenth
  Workshop on Innovative Use of NLP for Building Educational Applications,
  Association for Computational Linguistics, Florence, Italy, pp 92--102,
  \doi{10.18653/v1/W19-4409},
  \urlprefix\url{https://www.aclweb.org/anthology/W19-4409}

\bibitem[{Ke and Ng(2019)}]{ijcai2019-879}
Ke Z, Ng V (2019) Automated essay scoring: A survey of the state of the art.
  In: Proceedings of the Twenty-Eighth International Joint Conference on
  Artificial Intelligence, {IJCAI-19}, International Joint Conferences on
  Artificial Intelligence Organization, pp 6300--6308,
  \doi{10.24963/ijcai.2019/879},
  \urlprefix\url{https://doi.org/10.24963/ijcai.2019/879}

\bibitem[{Kenyon and Tschirner(2000)}]{kenyon2000rating}
Kenyon DM, Tschirner E (2000) The rating of direct and semi-direct oral
  proficiency interviews: Comparing performance at lower proficiency levels.
  The Modern Language Journal 84(1):85--101

\bibitem[{Kumar et~al.(2019)Kumar, Aggarwal, Mahata, Shah, Kumaraguru, and
  Zimmermann}]{kumar2019get}
Kumar Y, Aggarwal S, Mahata D, Shah RR, Kumaraguru P, Zimmermann R (2019) Get
  it scored using autosas—an automated system for scoring short answers. In:
  Proceedings of the AAAI Conference on Artificial Intelligence, vol~33, pp
  9662--9669

\bibitem[{Kumar et~al.(2020)Kumar, Bhatia, Kabra, Li, Jin, and
  Shah}]{kumar2020calling}
Kumar Y, Bhatia M, Kabra A, Li JJ, Jin D, Shah RR (2020) Calling out bluff:
  Attacking the robustness of automatic scoring systems with simple adversarial
  testing. arXiv preprint arXiv:200706796

\bibitem[{Loukina et~al.(2015)Loukina, Zechner, Chen, and
  Heilman}]{loukina-etal-2015-feature}
Loukina A, Zechner K, Chen L, Heilman M (2015) Feature selection for automated
  speech scoring. In: Proceedings of the Tenth Workshop on Innovative Use of
  {NLP} for Building Educational Applications, Association for Computational
  Linguistics, Denver, Colorado, pp 12--19, \doi{10.3115/v1/W15-0602},
  \urlprefix\url{https://www.aclweb.org/anthology/W15-0602}

\bibitem[{Loukina et~al.(2017)Loukina, Madnani, and
  Cahill}]{loukina-etal-2017-speech}
Loukina A, Madnani N, Cahill A (2017) Speech- and text-driven features for
  automated scoring of {E}nglish speaking tasks. In: Proceedings of the
  Workshop on Speech-Centric Natural Language Processing, Association for
  Computational Linguistics, Copenhagen, Denmark, pp 67--77,
  \doi{10.18653/v1/W17-4609},
  \urlprefix\url{https://www.aclweb.org/anthology/W17-4609}

\bibitem[{Lu(2010)}]{lu}
Lu X (2010) Automatic analysis of syntactic complexity in second language
  writing. International Journal of Corpus Linguistics 15:474--496,
  \doi{10.1075/ijcl.15.4.02lu}

\bibitem[{Lundberg and Lee(2017)}]{shap}
Lundberg SM, Lee SI (2017) A unified approach to interpreting model
  predictions. In: Guyon I, Luxburg UV, Bengio S, Wallach H, Fergus R,
  Vishwanathan S, Garnett R (eds) Advances in Neural Information Processing
  Systems 30, Curran Associates, Inc., pp 4765--4774,
  \urlprefix\url{http://papers.nips.cc/paper/7062-a-unified-approach-to-interpreting-model-predictions.pdf}

\bibitem[{Maddela and Xu(2018)}]{wordlist}
Maddela M, Xu W (2018) A word-complexity lexicon and a neural readability
  ranking model for lexical simplification. In: Proceedings of the Conference
  on Empirical Methods in Natural Language Processing (EMNLP)

\bibitem[{Malinin et~al.(2017)Malinin, Knill, Ragni, Wang, and
  Gales}]{Malinin2017}
Malinin A, Knill K, Ragni A, Wang Y, Gales M (2017) An attention based model
  for off-topic spontaneous spoken response detection: An initial study. In:
  Proc. 7th ISCA Workshop on Speech and Language Technology in Education, pp
  144--149, \doi{10.21437/SLaTE.2017-25},
  \urlprefix\url{http://dx.doi.org/10.21437/SLaTE.2017-25}

\bibitem[{Malone(2000)}]{malone2000simulated}
Malone M (2000) Simulated oral proficiency interviews: Recent developments.
  eric digest.

\bibitem[{McAuliffe et~al.(2017)McAuliffe, Socolof, Mihuc, Wagner, and
  Sonderegger}]{mcauliffe2017montreal}
McAuliffe M, Socolof M, Mihuc S, Wagner M, Sonderegger M (2017) Montreal forced
  aligner: Trainable text-speech alignment using kaldi. In: Interspeech, vol
  2017, pp 498--502

\bibitem[{McFee et~al.(2015)McFee, Raffel, Liang, Ellis, McVicar, Battenberg,
  and Nieto}]{librosa}
McFee B, Raffel C, Liang D, Ellis DP, McVicar M, Battenberg E, Nieto O (2015)
  librosa: Audio and music signal analysis in python. In: Proceedings of the
  14th python in science conference, vol~8, pp 18--25

\bibitem[{Merrick and Taly(2020)}]{shape}
Merrick L, Taly A (2020) The explanation game: Explaining machine learning
  models using shapley values. In: Holzinger A, Kieseberg P, Tjoa AM, Weippl E
  (eds) Machine Learning and Knowledge Extraction, Springer International
  Publishing, Cham, pp 17--38

\bibitem[{Molnar et~al.(2020)Molnar, König, Bischl, and
  Casalicchio}]{pdpaplication}
Molnar C, König G, Bischl B, Casalicchio G (2020) Model-agnostic feature
  importance and effects with dependent features -- a conditional subgroup
  approach

\bibitem[{Page(1966)}]{page1966imminence}
Page EB (1966) The imminence of... grading essays by computer. The Phi Delta
  Kappan 47(5):238--243

\bibitem[{{Panayotov} et~al.(2015){Panayotov}, {Chen}, {Povey}, and
  {Khudanpur}}]{7178964}
{Panayotov} V, {Chen} G, {Povey} D, {Khudanpur} S (2015) Librispeech: An asr
  corpus based on public domain audio books. In: 2015 IEEE International
  Conference on Acoustics, Speech and Signal Processing (ICASSP), pp 5206--5210

\bibitem[{Parekh et~al.(2020)Parekh, Singla, Chen, Li, and Shah}]{parekh2020my}
Parekh S, Singla YK, Chen C, Li JJ, Shah RR (2020) My teacher thinks the world
  is flat! interpreting automatic essay scoring mechanism. arXiv preprint
  arXiv:201213872

\bibitem[{Patil et~al.(2020)Patil, Singla, Shah, Hama, and
  Zimmermann}]{patil2020towards}
Patil R, Singla YK, Shah RR, Hama M, Zimmermann R (2020) Towards modelling
  coherence in spoken discourse. arXiv preprint arXiv:210100056
  \urlprefix\url{https://arxiv.org/abs/2101.00056}

\bibitem[{{Qian} et~al.(2018){Qian}, {Ubale}, {Mulholland}, {Evanini}, and
  {Wang}}]{8639697}
{Qian} Y, {Ubale} R, {Mulholland} M, {Evanini} K, {Wang} X (2018) A
  prompt-aware neural network approach to content-based scoring of non-native
  spontaneous speech. In: 2018 IEEE Spoken Language Technology Workshop (SLT),
  pp 979--986

\bibitem[{{Qian} et~al.(2019){Qian}, {Lange}, {Evanini}, {Pugh}, {Ubale},
  {Mulholland}, and {Wang}}]{8683717}
{Qian} Y, {Lange} P, {Evanini} K, {Pugh} R, {Ubale} R, {Mulholland} M, {Wang} X
  (2019) Neural approaches to automated speech scoring of monologue and
  dialogue responses. In: ICASSP 2019 - 2019 IEEE International Conference on
  Acoustics, Speech and Signal Processing (ICASSP), pp 8112--8116

\bibitem[{Raina et~al.(2020)Raina, Gales, and
  Knill}]{raina-etal-2020-complementary}
Raina V, Gales M, Knill K (2020) Complementary systems for off-topic spoken
  response detection. In: Proceedings of the Fifteenth Workshop on Innovative
  Use of NLP for Building Educational Applications, Association for
  Computational Linguistics, Seattle, WA, USA â†’ Online, pp 41--51,
  \doi{10.18653/v1/2020.bea-1.4},
  \urlprefix\url{https://www.aclweb.org/anthology/2020.bea-1.4}

\bibitem[{Riordan et~al.(2017)Riordan, Horbach, Cahill, Zesch, and
  Lee}]{riordan-etal-2017-investigating}
Riordan B, Horbach A, Cahill A, Zesch T, Lee CM (2017) Investigating neural
  architectures for short answer scoring. In: Proceedings of the 12th Workshop
  on Innovative Use of {NLP} for Building Educational Applications, Association
  for Computational Linguistics, Copenhagen, Denmark, pp 159--168,
  \doi{10.18653/v1/W17-5017},
  \urlprefix\url{https://www.aclweb.org/anthology/W17-5017}

\bibitem[{Shah et~al.(2021)Shah, Singla, Chen, and Shah}]{shah2021all}
Shah J, Singla YK, Chen C, Shah RR (2021) What all do audio transformer models
  hear? probing acoustic representations for language delivery and its
  structure. arXiv preprint arXiv:210100387

\bibitem[{Shapley(1953)}]{shapth}
Shapley LS (1953) A value for n-person games. Contributions to the Theory of
  Games 2(28):307--317

\bibitem[{Shashidhar et~al.(2015)Shashidhar, Pandey, and
  Aggarwal}]{aspiringminds}
Shashidhar V, Pandey N, Aggarwal V (2015) Automatic spontaneous speech grading:
  A novel feature derivation technique using the crowd. In: Proceedings of the
  53rd Annual Meeting of the Association for Computational Linguistics and the
  7th International Joint Conference on Natural Language Processing (Volume 1:
  Long Papers), Association for Computational Linguistics, Beijing, China, pp
  1085--1094, \doi{10.3115/v1/P15-1105},
  \urlprefix\url{https://www.aclweb.org/anthology/P15-1105}

\bibitem[{Singla et~al.(2021{\natexlab{a}})Singla, Gupta, Bagga, Chen,
  Krishnamurthy, and Shah}]{singla2021speaker}
Singla YK, Gupta A, Bagga S, Chen C, Krishnamurthy B, Shah RR
  (2021{\natexlab{a}}) Speaker-conditioned hierarchical modeling for automated
  speech scoring. In: Proceedings of the 30th ACM International Conference on
  Information \& Knowledge Management, pp 1681--1691

\bibitem[{Singla et~al.(2021{\natexlab{b}})Singla, Krishna, Shah, and
  Chen}]{singla2021using}
Singla YK, Krishna S, Shah RR, Chen C (2021{\natexlab{b}}) Using sampling to
  estimate and improve performance of automated scoring systems with
  guarantees. arXiv preprint arXiv:211108906

\bibitem[{Singla et~al.(2021{\natexlab{c}})Singla, Parekh, Singh, Li, Shah, and
  Chen}]{singla2021aes}
Singla YK, Parekh S, Singh S, Li JJ, Shah RR, Chen C (2021{\natexlab{c}}) Aes
  systems are both overstable and oversensitive: Explaining why and proposing
  defenses. arXiv preprint arXiv:210911728

\bibitem[{Stansfield and Winke(2008)}]{stansfield2008testing}
Stansfield C, Winke P (2008) Testing aptitude for second language learning.
  Encyclopaedia of language and education, 2nd Edition: Language Testing and
  assessment 7:81--94

\bibitem[{Taghipour and Ng(2016)}]{taghipour-ng-2016-neural}
Taghipour K, Ng HT (2016) A neural approach to automated essay scoring. In:
  Proceedings of the 2016 Conference on Empirical Methods in Natural Language
  Processing, Association for Computational Linguistics, Austin, Texas, pp
  1882--1891, \doi{10.18653/v1/D16-1193},
  \urlprefix\url{https://www.aclweb.org/anthology/D16-1193}

\bibitem[{Tao et~al.(2014)Tao, Evanini, and Wang}]{7078590}
Tao J, Evanini K, Wang X (2014) The influence of automatic speech recognition
  accuracy on the performance of an automated speech assessment system. In:
  2014 IEEE Spoken Language Technology Workshop (SLT), pp 294--299,
  \doi{10.1109/SLT.2014.7078590}

\bibitem[{Tay et~al.(2018)Tay, Phan, Tuan, and Hui}]{AAAI1816431}
Tay Y, Phan M, Tuan LA, Hui SC (2018) Skipflow: Incorporating neural coherence
  features for end-to-end automatic text scoring. In: AAAI Conference on
  Artificial Intelligence

\bibitem[{Tilk and Alum{\"a}e(2016)}]{tilk2016}
Tilk O, Alum{\"a}e T (2016) Bidirectional recurrent neural network with
  attention mechanism for punctuation restoration. In: Interspeech 2016

\bibitem[{Wang and Evanini(2019)}]{chcontent}
Wang X, Evanini K (2019) Features measuring content and discourse coherence.
  Automated Speaking Assessment: Using Language Technologies to Score
  Spontaneous Speech p 138

\bibitem[{Xi et~al.(2008)Xi, Higgins, Zechner, and
  Williamson}]{xi2008automated}
Xi X, Higgins D, Zechner K, Williamson DM (2008) Automated scoring of
  spontaneous speech using speechratersm v1. 0. ETS Research Report Series
  2008(2):i--102

\bibitem[{Yoon and Lee(2019)}]{yoon-lee-2019-content}
Yoon SY, Lee CM (2019) Content modeling for automated oral proficiency scoring
  system. In: Proceedings of the Fourteenth Workshop on Innovative Use of NLP
  for Building Educational Applications, Association for Computational
  Linguistics, Florence, Italy, pp 394--401, \doi{10.18653/v1/W19-4441},
  \urlprefix\url{https://www.aclweb.org/anthology/W19-4441}

\bibitem[{Yoon et~al.(2018)Yoon, Loukina, Lee, Mulholland, Wang, and
  Choi}]{yoon-etal-2018-word}
Yoon SY, Loukina A, Lee CM, Mulholland M, Wang X, Choi I (2018) Word-embedding
  based content features for automated oral proficiency scoring. In:
  Proceedings of the Third Workshop on Semantic Deep Learning, Association for
  Computational Linguistics, Santa Fe, New Mexico, pp 12--22,
  \urlprefix\url{https://www.aclweb.org/anthology/W18-4002}

\bibitem[{{Yu} et~al.(2015){Yu}, {Ramanarayanan}, {Suendermann-Oeft}, {Wang},
  {Zechner}, {Chen}, {Tao}, {Ivanou}, and {Qian}}]{7404814}
{Yu} Z, {Ramanarayanan} V, {Suendermann-Oeft} D, {Wang} X, {Zechner} K, {Chen}
  L, {Tao} J, {Ivanou} A, {Qian} Y (2015) Using bidirectional lstm recurrent
  neural networks to learn high-level abstractions of sequential features for
  automated scoring of non-native spontaneous speech. In: 2015 IEEE Workshop on
  Automatic Speech Recognition and Understanding (ASRU), pp 338--345

\bibitem[{Zhang et~al.(2018)Zhang, Geiger, Pohjalainen, Mousa, Jin, and
  Schuller}]{10.1145/3178115}
Zhang Z, Geiger J, Pohjalainen J, Mousa AED, Jin W, Schuller B (2018) Deep
  learning for environmentally robust speech recognition: An overview of recent
  developments. ACM Trans Intell Syst Technol 9(5), \doi{10.1145/3178115},
  \urlprefix\url{https://doi.org/10.1145/3178115}

\end{thebibliography}


\end{document}